\documentclass[10pt,twocolumn,letterpaper]{article}

\usepackage[pagenumbers]{cvpr} 

\usepackage{xcolor}
\definecolor{darkgreen}{RGB}{0,100,0}
\definecolor{mediumgreen}{RGB}{0,128,0}
\definecolor{lightgreen}{RGB}{144,238,144}
\definecolor{limegreen}{RGB}{50,205,50}
\definecolor{forestgreen}{RGB}{34,139,34}

\newcommand{\todel}[1]{{\color{gray} \em (#1)}}

\definecolor{cvprblue}{rgb}{0.21,0.49,0.74}
\usepackage[pagebackref,breaklinks,colorlinks,allcolors=cvprblue]{hyperref}
\usepackage{amsmath,epstopdf,amssymb,color,url,multirow,multicol,verbatim,threeparttable,diagbox,makecell,booktabs,colortbl}
\usepackage{algorithm,algorithmic,pifont}

\newcommand{\tft}{\textbf}

\newcommand{\cblue}{\color{blue}}

\definecolor{light-gray}{gray}{0.82}
\definecolor{lightblue}{RGB}{100, 149, 237}
\definecolor{lightgreen}{RGB}{60, 179, 113}

\newcommand{\Ours}{ForenX}
\newcommand{\Ourdata}{ForgReason}

\title{ForenX: Towards Explainable AI-Generated Image Detection with Multimodal Large Language Models}

\author{
Chuangchuang Tan$^{1,2}$\thanks{Work done during internship at MSRA.}\quad
Jinglu Wang$^{2}$\quad 
Xiang Ming$^{2}$\quad
Renshuai Tao$^{1}$\quad
Yunchao Wei$^{1}$\quad\\
Yao Zhao$^{1}$\quad
Yan Lu$^{2}$\quad\\[1.2mm]
$^1$Beijing Jiaotong University \quad
$^2$Microsoft Research Asia
}

\begin{document}

\twocolumn[{%
\renewcommand\twocolumn[1][]{#1}%
\maketitle

\begin{center}
  \centering
  \vspace{-0.95cm}
  \captionsetup{type=figure}
  \includegraphics[width=1.0\textwidth]{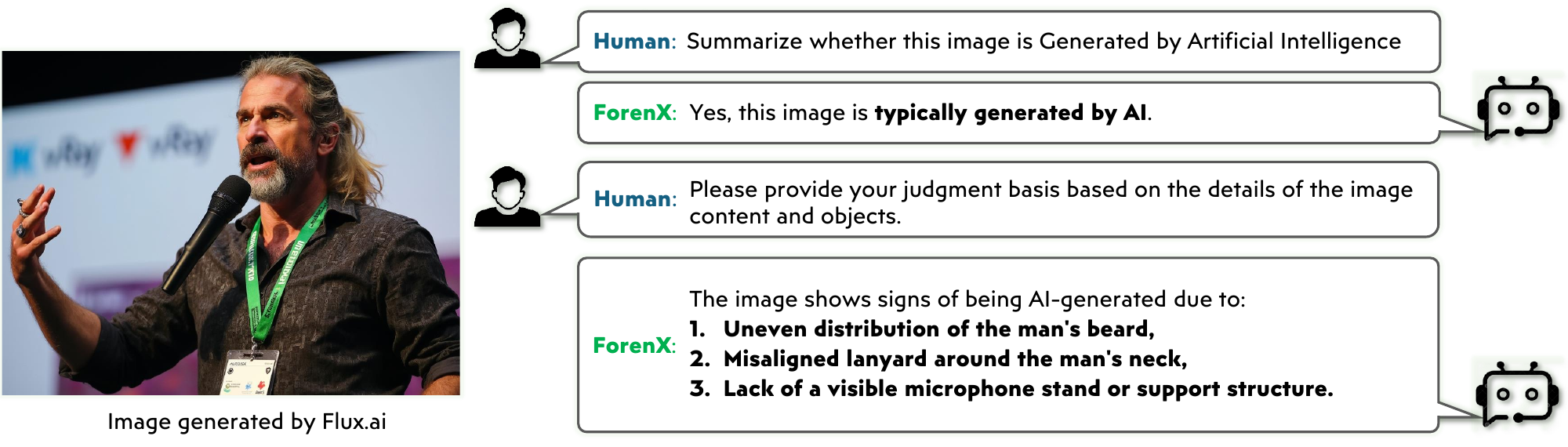}
  \caption{
  We present a sample result of our explainable AI-generated image detection method, \Ours. Given an image created by Flux.ai \cite{Flux}, \Ours~not only delivers the detection result (``Yes''), but also offers a forensic analysis that resonates with human reasoning. For instance, the third evidence, highlighting the lack of a supporting structure for the microphone, is closely aligned with human intuition, which standard MLLMs fail to provide (demonstrated in our experiments).
  }
  \label{fig:teaser}
\end{center}
\vspace{0.95em}
}]
\renewcommand{\thefootnote}{\fnsymbol{footnote}}  
\setcounter{footnote}{0}  
\footnotetext[1]{ Work done during C. Tan's internship at MSRA.}
\renewcommand{\thefootnote}{\arabic{footnote}}
\setcounter{footnote}{0} 

\begin{abstract}

\vspace{-1.3em}
Advances in generative models have led to AI-generated images visually indistinguishable from authentic ones. Despite numerous studies on detecting AI-generated images with classifiers, a gap persists between such methods and human cognitive forensic analysis. We present \Ours, a novel method that not only identifies the authenticity of images but also provides explanations that resonate with human thoughts. \Ours~employs the powerful multimodal large language models (MLLMs) to analyze and interpret forensic cues. Furthermore, we overcome the limitations of standard MLLMs in detecting forgeries by incorporating a specialized \textbf{forensic prompt} that directs the MLLMs' attention to forgery-indicative attributes. This approach not only enhance the generalization of forgery detection but also empowers the MLLMs to provide explanations that are accurate, relevant, and comprehensive. Additionally, we introduce \textbf{\Ourdata}, a dataset dedicated to descriptions of forgery evidences in AI-generated images. 
Curated through collaboration between an LLM-based agent and a team of human annotators, this process provides refined data that further enhances our model’s performance.
We demonstrate that even limited manual annotations significantly improve explanation quality. 
{We evaluate the effectiveness of \Ours~ on two major benchmarks. }
The model's explainability is verified by comprehensive subjective evaluations.

\end{abstract}    
\section{Introduction}
\label{sec:intro}

\begin{figure*}[ht!]
  \centering
   \includegraphics[width=\textwidth]{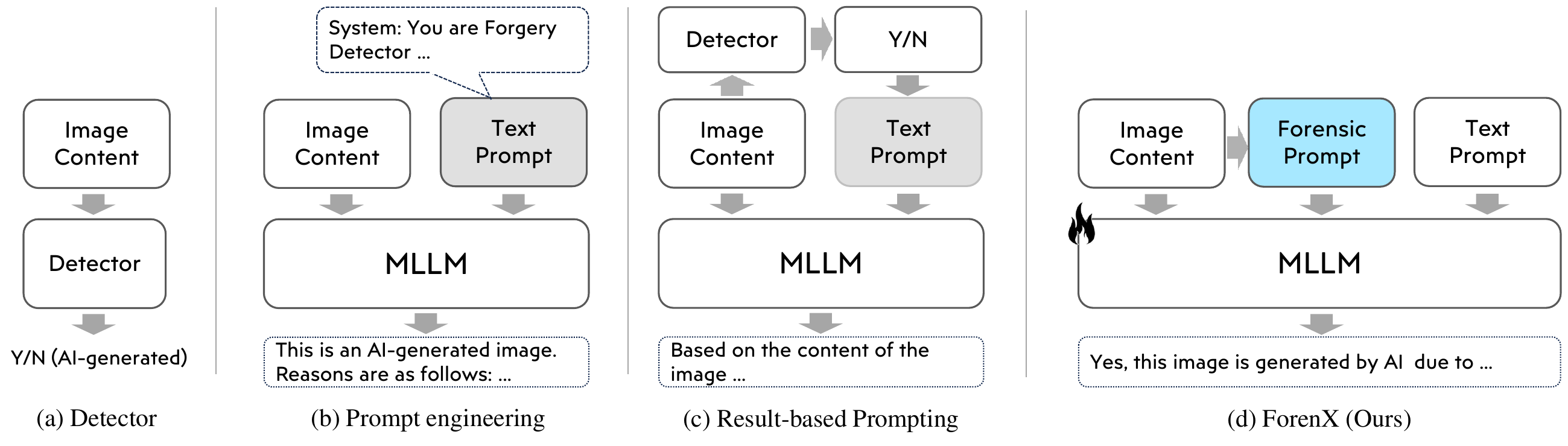}
  \caption{Empower MLLMs to detect AI-generated Images with reliable explanations. Detection model (a) only outputs binary classification without human-understood explanations. Prompt engineering (b), even integrated with detection result (c) may generate explanations without focusing on forensic evidences. Our method (d) enhances the MLLM by incorporating the specialized forensic prompt.}
  \label{fig:intro}
\end{figure*}

Recent advancements in generative models, such as GANs and diffusion models \cite{goodfellow2014generative, karras2018progressive, karras2019style, ho2020denoising, rombach2022high}, have led to the creation of AI-generated images that are nearly indistinguishable from real-world photographs, raising concerns about potential misuse. As a result, developing effective methods for detecting AI-generated images has become a critical research focus.

Recent studies in AI-generated image detection \cite{Tan2023CVPR, Durall, tan2024rethinking, liu2024forgery, chendrct,liu2024evolving} have predominantly focused on detection-based approaches (\cref{fig:intro} (a)), aiming to differentiate AI-generated and authentic images by detecting specific artifacts or patterns introduced by generative models. These methods rely on artifact-based features to capture subtle traces of forgery.
However, they are prone to the classic overfitting issue, restricting their capacity to generalize across new generative models.

Moreover, detection models, based on binary classification, typically identify indicative patterns in a black-box manner. Such patterns, often imperceptible or incomprehensible to humans, resulting in difficulties for humans to verify their detection results.
In contrast, human observers spot AI-generated images by examining not only artifacts but also the content, such as unnatural human poses, inconsistent lighting conditions, or violations of physical laws. There is a significant discrepancy in the forensic analysis by detection models and humans.
To bridge this gap, 
  we introduce an approach that goes beyond binary classification.
Our method enriches the forensic analysis of detection results, delivering explanations that highlight both content-related anomalies and generation-based artifacts, thereby aligning the model closer to human perceptual and cognitive strategies. 
We term our method as the \textbf{Foren}sics e\textbf{X}plainer (\textbf{\Ours}).
Additionally, model-generated explanations can improve user understanding of the detection results and facilitate a “double-check” process, enabling human verification of the model's predictions.

Multimodal Large Language Models (MLLMs), such as LLava~\cite{llava} and GPT-4 Vision~\cite{achiam2023gpt}, demonstrate compelling abilities in various cross-modal tasks, including understanding and reasoning. These abilities render them highly advantageous for achieving explainable AI-generated image detection.
Adopting prompt engineering in the straightforward manner to guide MLLM outputs (\cref{fig:intro} (b)) tends to generate hallucinations, which may not provide meaningful explanations.
Even when prompts are integrated with predetermined classification results (\cref{fig:intro} (c)), the approach proves unsatisfactory. 
Likewise, enhancing MLLMs by simply fine-tuning them with textual annotations does not suffice for reliable explanation.
All the effectiveness is attributed to the inherent lack of considerations for image authenticity\footnote{
    Authenticity in this paper refers to the likelihood an image is captured in the real world rather than generated by AI.
  } within the standard MLLMs.
To overcome this limitation,
  we introduce the \textit{\textbf{forensic prompt}}, a specialized input designed to direct MLLMs towards prioritizing authenticity and its related cues, which are beyond conventional text and image inputs.
This strategy, integrated with LoRA~\cite{hu2021lora} fine-tuning, enhances MLLMs' ability to achieve both robust detection and reliable explanation generation.

The forensic prompt is considered to fulfill three essential requirements for effectiveness.
(1) Initially, it should enable multi-modal representation with alignment of text and image features.
We derive inspiration from the detector~\cite{ojha2023towards}, which showcases the utility of pretrained CLIP-ViT~\cite{dosovitskiy2020image,
  radford2021learning} features. We construct our forensic prompt based on the CLIP-ViT features. However, we observe that the pretrained features, designed for general purpose, may not inherently possess the specific authenticity. 
(2) The prompt should encapsulate effective cues indicative of authenticity.
We integrate the features with authenticity attributes by tuning them specifically for detection with an additional detection loss. 
(3) Finally, the prompt should be adaptable to MLLMs. To ensure this, we fine-tune these enhanced features along with the forensic prompt during the MLLM training phase. This process is aimed at intensifying MLLMs' focus on authenticity cues, making them more adept at identifying authentic content.
As demonstrated in our experiment,
our model with fine-tuned CLIP-ViT features achieves significantly better performance compared to the untuned ones.

Since no existing dataset provides annotations detailing why an image appears AI-generated, 
  we initially train our model using captions generated by the method in~\cref{fig:intro}(b), 
  as it is impractical to obtain large-scale annotations from human annotators for AI-generated images.
However, these machine-generated captions lack alignment with human reasoning. 
To address this, 
    we collected human-provided annotations specifying the attributes that make an image seem AI-generated.
We then fine-tune the model,
  initially trained on machine generated low-quality captions,
  using this human-annotated data.   
This approach allows the model to attain human-like reasoning capabilities with a minimal set of human annotations.

In conclusion, we present \Ours, 
the first approach to extend MLLMs for Explainable AI-generated Image Detection. 
  Our contributions are three-fold:

\begin{itemize}

\item We propose a simple yet effective pipeline, \Ours, to enable MLLMs for explainable AI-generated image detection. \Ours~demonstrates strong recognition capabilities, generalization performance, and explainability.

\item We introduce the concept of \textbf{\textit{forensic prompt}} to guide MLLMs, enhancing their ability to detect AI-generated images accurately.

\item We develop an explainable AI-generated image detection dataset, \textbf{\textit{ForgReason}}, which pairs images with authenticity-related descriptions annotated by human annotators and GPT-4 Vision, facilitating alignment with human reasoning.

\end{itemize}
\section{Related Works}

\paragraph{AI-Generated image detection.}
To improve the generalizability of models in AI-generated image detection, current research primarily focuses on artifact extraction and detector design within this domain. Various low-level artifact features have been introduced to capture generation traces, including frequency-based attributes \cite{jeong2022bihpf}, gradient information \cite{Tan_2023_CVPR}, neighboring pixel relationships \cite{tan2024rethinking}, and random-mapping features \cite{tan2024data}. For example, BiHPF \cite{jeong2022bihpf} enhances artifact magnitudes using dual high-pass filters, while LGrad \cite{Tan_2023_CVPR} utilizes gradient data from pre-trained models as representations of artifacts. NPR \cite{tan2024rethinking} presents a simple yet effective approach by reconsidering up-sampling operations for artifact representation.
In addition to low-level features, large pre-trained models have been leveraged to capture high-level forgery traces in AI-generated content detection tasks. UniFD \cite{ojha2023towards} directly incorporates image features from the CLIP model for linear classification, showcasing robust deepfake detection even with unseen sources. FatFormer \cite{liu2024forgery} combines frequency analysis with a text encoder as an adapter to the frozen CLIP vision model, thereby enhancing detection performance significantly. After investigating the mechanism of CLIP in deepfake detection, C2P-CLIP\cite{tan2024c2p} introduces the category common prompt to enhance detection accuracy.

\paragraph{Multimodal large language model.}
In recent years, the advent of Multimodal Large Language Models (MLLMs) such as GPT-4V~\cite{achiam2023gpt} and LLaVA~\cite{liu2024visual} has garnered significant attention due to their unparalleled proficiency in image comprehension and analysis. 
Several studies~\cite{jia2024can,shi2024shield,NEURIPS2022_9d560961} have employed prompt engineering to explore the applicability of MLLMs on new
tasks, such as face forgery analysis and reasoning.
Moreover, research efforts~\cite{dai2023instructblip,li2023mimic,li2024llava} have been directed towards instruction tuning to augment the performance of large language models on specialized tasks. 
Recent advancements in deepfake detection have introduced methodologies leveraging MLLM.
SNIFFER \cite{qi2024sniffer} employs InstructBLIP \cite{dai2023instructblip} for the enhancement of misinformation detection capabilities, showcasing its significant contributions to the domain of information security.
FFAA~\cite{huang2024ffaa} integrates a fine-tuned Multimodal LLM with a Multi-answer Intelligent Decision System for Open-World Face Forgery Analysis VQA. 
ForgeryGPT~\cite{li2024forgerygpt} utilizes precise forgery mask data to refine LLM capabilities for explainable image forgery detection. 
Jia \etal \cite{jia2024can} meticulously crafted prompts to assess MLLM's effectiveness in face forgery detection.
Despite initial explorations into leveraging MLLM for detecting forgery image, existing studies have mainly focused on identifying facial manipulations and exhibit a gap in AIGC forgery detection.

\section{Method}
 \begin{figure*}[ht!]
 \vspace{-0.45 cm}
   \centering
    \includegraphics[width=1.0\textwidth]{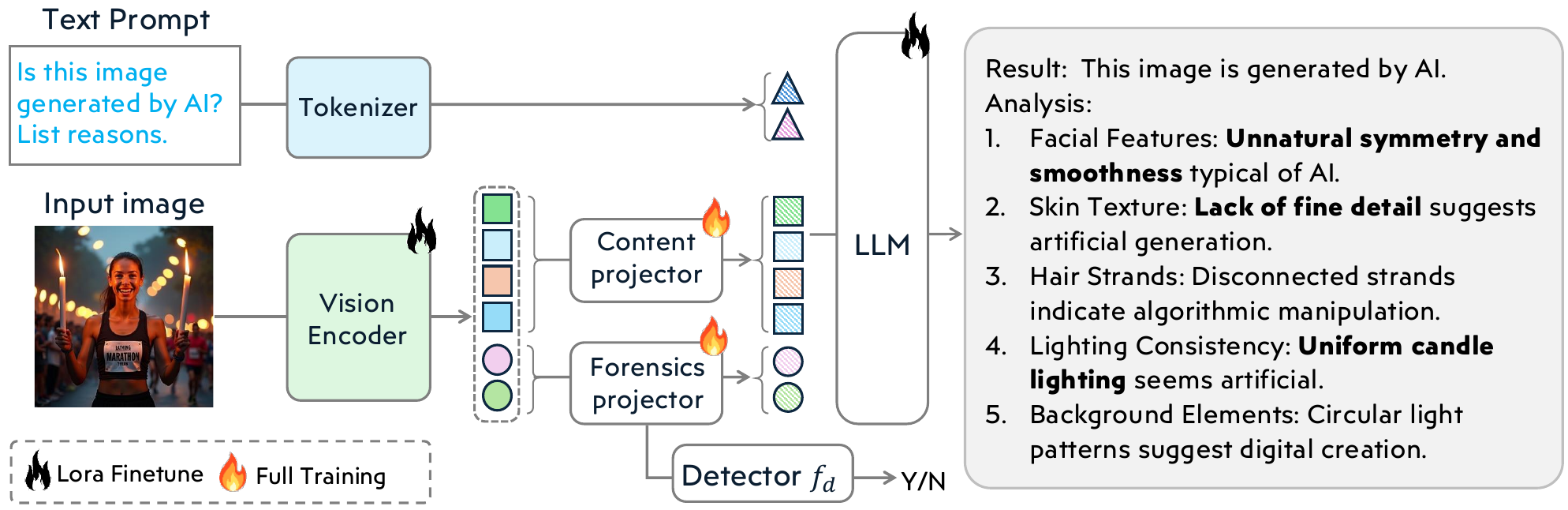}
    \caption{Overview of our ForenX: A Simple Yet Effective Explainable AI-Generated Image Detection Pipeline. To overcome the limitations of standard MLLMs in detecting forgeries, we incorporate a specialized forensic prompt that directs the MLLMs’ attention to forgery-indicative attributes.}
    \label{fig:pipelines}
    \vspace{-0.45 cm}
 \end{figure*}

In this work, we aim to develop a Multimodal Large Language Model (MLLM) capable of detecting AI-generated images, with a focus on three key attributes: recognition ability, generalizability, and explainability. An overview of the \Ours~architecture is shown in \cref{fig:pipelines}. Our primary approach involves incorporating a forensic prompt into the MLLM, such as LLava\cite{llava}. In the remainder of this section, we describe the technical aspects of \Ours, emphasizing the construction of the forensic prompt and the two-stage training strategy. We also outline the creation of the \textit{ForenReason} dataset.

\if false
\todel{
In this work, we aim to develop a multimodal large language model (MLLM) capable of detecting AI-generated images, focusing on three key attributes: {recognition ability}, {generalizibility}, and {explainability}. 
The pipeline overview is illustrated in \cref{fig:pipelines}. Our primary strategy involves integrating forensics semantic information into the MLLM, such as LLava. Specifically, unlike traditional MLLMs that only extract image content from the visual encoder, our approach extracts both content semantics and forensics semantics. Subsequently, we introduce these two types of semantic information into the MLLM to equip it with enhanced AI-generated images detection capabilities.
}
\fi

\subsection{The Architecture of \Ours}

As discussed in~\cref{sec:intro},
  we introduce an forensic prompt to guide the MLLM's focus on assessing the authenticity of the image.
Leveraging the strong performance of CLIP-ViT~\cite{dosovitskiy2020image} for this task,
  we use features extracted from CLIP-ViT to construct the forensic prompt.
To maintain feature consistency and reduce training instability,
  we also use CLIP-ViT features as the representation of image content, 
  ensuring aligned feature distributions across inputs.

\if false
\todel{
Formally, the construction of the forensic prompts is as follows:
\begin{equation}
  \begin{aligned}
      F_v &= g(X_v),\\
      F_v^f &= l_f(F_v),\\
      H_v^f &= m_f(F_v^f),
  \end{aligned}
  \label{eq:eq1}
\end{equation}
in this setup, 
  $X_v$ denotes the input image, 
  $g$ represents the feature extractor and 
  $F_v$ refers to the extracted features.
We then apply forensics projector $l_f$ 
  to project these features into the forensics embedding space,
  resulting in $F_v^f$,
  which also serves as the input for an auxiliary forgery detection loss,
  providing additional constraints on the forensic embedding space.
Here $H_v^f$ serves as the forensic prompt in our model and 
  $m_f$ is the mapping function from the forensics embedding space to the word embedding space.
}
\fi

Formally, the construction of the forensic prompts is as follows:
\begin{equation}
  \begin{aligned}
      F_v &= g(X_v),\\
      F_v^f &= l_f(F_v),\\
      H_v^f &= m_f(F_v^f),
  \end{aligned}
  \label{eq:eq1}
\end{equation}
in this setup, 
  $X_v$ denotes the input image, 
  $g$ represents the feature extractor and 
  $F_v$ refers to the extracted features. 
  The $l_f$ and $m_f$ constitute the forensic projector designed to extract forensic prompt.
We apply forensics encoder $l_f$ 
  to project these features into the forensics embedding space,
  resulting in $F_v^f$,
  which also serves as the input for an auxiliary forgery detection loss,
  providing additional constraints on the forensic embedding space.
Here $H_v^f$ serves as the forensic prompt in our model and 
  $m_f$ is the mapping function from the forensics embedding space to the word embedding space.

We use a trainable forensics embedding $d$ as the forensics encoder $l_f$ in our experiments. 
Specifically, the forensics embedding $d$ transforms the visual feature $F_v$ into forensics features $F_v^f$ as follows:
 \begin{equation}
\begin{split}
    F_v^f  = F_v \otimes d .
 \end{split}
  \label{eq:eq3}
\end{equation}
In our experiments, $\otimes$ denotes the Hadamard product. 
To further integrate forensics information into $F_v^f$, 
  we impose constraints using forgery detection labels $y_d \in \{0, 1\}$:
\begin{equation}
  \begin{split}
    \mathcal{L}_{detection} = Loss(f_d(F_v^f), y_d),
  \end{split}
  \label{eq:eq3_2}
\end{equation}
in which $f_d$ is a function that maps feature $F_v^f$ to the classification space. 
In our experiment, we we employ a simple summation function $sum(\cdot)$ as $f_d$. 
We then connect $F_v^f$ into the word embedding space using $m_f$. This mapping is implemented using two MLP layers.

To achieve comprehensive content understanding, AI-generated image detection and forensics analysis, we feed the text tokens $H_t$ (from text prompts), visual content tokens $H_v^c$(same as LLava), and forensic prompt $H_v^f$ into the Large Language Model(LLM), as follows:
\begin{equation}
  \begin{split}
    T_p = llm(H_t, H_v^c, H_v^f). \\
  \end{split}
  \label{eq:eq4}
\end{equation}
We utilize conversations with image captions, detection results, and forensics descriptions as labels for instruction fine-tuning of LLM:
\begin{equation}
\begin{split}
  \mathcal{L}_{instruction} = Loss(T_p, T_l). \\
 \end{split}
  \label{eq:eq5}
\end{equation}
in which $T_l$ is the conversational data for instruction tuning.
Following the foundation MLLM, 
  we adopt instruction-tuning using an auto-regressive objective, 
  enabling model to predict subsequent outputs based on input instructions. 

In our pipeline, 
  it is essential that the LLM can independently identify AI-generated images to generate convincing forensic reasoning.
To achieve this, 
  we avoid directly converting detection results into prompts (e.g., “this image is real” or “this image is AI-generated”).
Instead, 
  we introduce a forensic prompt to guide the LLM's attention to both the semantic content and authenticity elements within the image, 
  prompting the model to generate more insightful forensic analyses.

\subsection{Dataset Construction}

To our knowledge, no existing dataset directly supports training MLLMs for AI-generated image detection, as we require image captions focused on authenticity. However, since modern AI-generated images closely mimic real images, creating large-scale labeled data through human annotation alone is impractical. A small sample size would also risk quick overfitting. To address this, we first use LLMs to generate a large set of preliminary captions focused on image authenticity. We then manually filter for cases that are difficult for state-of-the-art binary classifiers but readily identifiable by human judgment. The data from LLMs serves as a pretraining set, while the human-labeled examples are used for fine-tuning, enhancing the model's alignment with human reasoning on image authenticity.

\subsubsection{Data Generation with LLM}
\label{sec:annotation_llm}

For fair comparison with other model in the paper,
  we only use the LLava~\cite{llava} to generate the image captions for two widely used forgery detection datasets, \textit{Genimage} and \textit{ForenSynths}.
For each image $X_v$ in the datasets,
  it is annotated with two-round conversational data $\{(X_q^1,X_a^1), (X_q^2,X_a^2)\}$. 
This includes two types of question-and-answer pairs: content-related $(X_q^1,X_a^1)$, and detection results$(X_q^2,X_a^2)$.

Specifically, responses related to image content are generated by the pre-trained LLava model, using questions that are randomly chosen from a predefined list within 
LLava\footnote{Some samples can be found in the supplementary materials.}
For questions about detection results, answers are directly derived from image labels. For example, if an image is labeled as AI-generated, the corresponding question and answer pair might be:

\setlength{\fboxrule}{0.5pt}
\vspace{0.3em}
\noindent
\fbox{\parbox{0.98\columnwidth}{
\begin{itemize}
\item \textbf{Question:} Summarize whether this image is generated by AI, please return beginning with \textbf{Yes} or \textbf{No}.
\item \textbf{Answer:} Yes, this image is typically generated by AI.
\end{itemize}}}
\vspace{0.01em}

The test samples will marked as AI-generated/authentic if the answer starts with \textbf{Yes}, and authentic if the answer starts with \textbf{Yes}/\textbf{No} 
  in the evaluation process.

\subsubsection{Data Annotation with Human}
\label{DataAnnotationwithHuman}
To further enhance the forensics interpretability of the LLM, 
  we involve human annotators to label 2,215 images with explanations on why each image is identified as AI-generated. 
The detailed annotation process is as follows:

\vspace{0.3em}
\noindent
\fbox{\parbox{0.98\columnwidth}{
\begin{enumerate}

    \item \textbf{Select Images:} Obtain images from Midjourney and choose those that have a realistic style.
    
    \item \textbf{Annotate with boxes}: Manually annotate areas in the images that appear unreasonable using box annotations. Provide a description explaining why these areas are deemed unreasonable.
    
    \item \textbf{Summarize with GPT-4 Vison:} Use GPT-4 Vision to summarize the images and the manual annotations, generating final explanations for why the images are considered AI-generated.

\end{enumerate}
}}
\vspace{0.01em}

These images were annotated with captions and detection results following the methods described in~\cref{sec:annotation_llm}. 
The detailed flowchart of Human-annotation is provided in the supplementary materials.
Before using GPT-4 Vision to generate summaries, we convert bounding box coordinates into textual descriptions that convey relative spatial positions. 
In this process, some detailed annotations are omitted to adhere to OpenAI's restrictions.

\subsection{Optimization Strategy}

Our training process follows a two-stage optimization strategy to address the imbalance between annotations obtained from the LLM and those from human annotators. Additionally, we introduce real-world images as negative samples in the second stage, as the initial 2,215 human-annotated images consist solely of AI-generated samples, which could lead to overfitting. 
We further sample 5000 real and 1000 fake from Genimage dataset, resulting in a total of 8,215 images.
Following is the optimization details for each stage:

\vspace{0.1em}
\noindent
\textbf{Stage 1: Initial training for general recognition and explainability}. 

We begin by using the pre-trained LLava model as the foundational MLLM, fine-tuning it on the ForenSynths and GenImage datasets, which provide content captions and detection labels. During this stage, both the CLIP-ViT encoder and the LLM are fine-tuned using LoRA. The overall loss function applied in this stage is:
\begin{equation}
\begin{split}
      \mathcal{L} = \mathcal{L}_{detection} + \mathcal{L}_{instruction}
 \end{split}
  \label{eq:eq6}
\end{equation}

\noindent
\textbf{Stage 2: Fine-tuning with the dedicated ForgReason.}. 
In this stage, 
  we utilize the 8,215 images to further enhance the explainability of the LLM. 
During this phase, 
  only the LLM is fine-tuned using LoRA with the instruction-based loss  
  $\mathcal{L}_{instruction}$, 
  while all other layers, including the CLIP-ViT, remain frozen. 
This fine-tuning process focuses on improving the model's ability to generate detailed and accurate forensics explanations, 
  guiding the model to better interpret and explain the authenticity of AI-generated images.

\section{Experiments}
\label{sec:experiments}

\begin{table*}[!ht]
\vspace{-0.5 cm}
    \centering
\resizebox{\textwidth}{35mm}{
    \begin{tabular}{l c c c c c c c c c c c c}
    \bottomrule \hline
        \multirow{2}*{\textbf{Methods}} & \multirow{2}*{\textbf{Venues}}  & \multicolumn{2}{c}{\textbf{Abilities}} & \multicolumn{9}{c}{\textbf{Test Models}}     \\
          \cmidrule(lr){3-4}\cmidrule(lr){5-13} ~ & ~ &    \textbf{Detection} & \textbf{Explain}  & \textbf{Midjourney} & \textbf{SDv1.4} & \textbf{SDv1.5} & \textbf{ADM} & \textbf{GLIDE} & \textbf{Wukong} & \textbf{VQDM} & \textbf{BigGAN} & \textbf{mAcc}\\
          \bottomrule \hline 
ResNet-50\cite{he2016deep}  &  CVPR2016   & \checkmark & $\times$ & 54.9 & 99.9 & 99.7 & 53.5 & 61.9 & 98.2 & 56.6 & 52.0 & 72.1 \\
DeiT-S\cite{touvron2021training}& ICML2021 & \checkmark & $\times$ & 55.6 & 99.9 & 99.8 & 49.8 & 58.1 & 98.9 & 56.9 & 53.5 & 71.6 \\
Swin-T\cite{liu2021swin}    &  ICCV2021   & \checkmark & $\times$ & 62.1 & 99.9 & 99.8 & 49.8 & 67.6 & 99.1 & 62.3 & 57.6 & 74.8 \\
CNNSpot\cite{wang2020cnn}    &  CVPR2020  & \checkmark & $\times$ & 52.8 & 96.3 & 95.9 & 50.1 & 39.8 & 78.6 & 53.4 & 46.8 & 64.2 \\
Spec\cite{zhang2019detecting} & WIFS2019  & \checkmark & $\times$ & 52.0 & 99.4 & 99.2 & 49.7 & 49.8 & 94.8 & 55.6 & 49.8 & 68.8 \\
F3Net\cite{qian2020thinking}  & ECCV2020  & \checkmark & $\times$ & 50.1 & 99.9 & 99.9 & 49.9 & 50.0 & 99.9 & 49.9 & 49.9 & 68.7 \\
GramNet\cite{liu2020global}   & CVPR2020  & \checkmark & $\times$ & 54.2 & 99.2 & 99.1 & 50.3 & 54.6 & 98.9 & 50.8 & 51.7 & 69.9 \\
UnivFD\cite{ojha2023towards}  & CVPR2023  & \checkmark & $\times$ & 93.9 & 96.4 & 96.2 & 71.9 & 85.4 & 94.3 & 81.6 & 90.5 & 88.8 \\
NPR \cite{tan2024rethinking}  &  CVPR2024 & \checkmark & $\times$ & 81.0 & 98.2 & 97.9 & 76.9 & 89.8 & 96.9 & 84.1 & 84.2 & 88.6 \\
FreqNet \cite{tan2024frequency} & AAAI2024 & \checkmark & $\times$ & 89.6 & 98.8 & 98.6 & 66.8 & 86.5 & 97.3 & 75.8 & 81.4 & 86.8 \\
FatFormer\cite{liu2024forgery} & CVPR2024 & \checkmark & $\times$ & 92.7 & 100.0 & 99.9 & 75.9 & 88.0 & 99.9 & 98.8 & 55.8 & 88.9 \\
DRCT\cite{chendrct} & ICML2024 & \checkmark & $\times$ & 91.5 & 95.0 & 94.4 & 79.4 & 89.2 & 94.7 & 90.0 & 81.7 & 89.5 \\
CLIP(336px)-Lora & - & \checkmark & $\times$ &  94.2 & 99.0 & 99.1 & 50.7 & 94.2 & 98.9 & 88.2 & 61.2 & 85.7  \\ 
\hline
LLAVA & NeurIPS2023 & \checkmark & \checkmark &  51.6 & 50.5 & 50.6 & 50.2 & 51.7 & 52.8 & 50.6 & 54.0 & 51.5  \\ 
LLAVA-PE & NeurIPS2023 & \checkmark & \checkmark &  50.6 & 50.4 & 50.4 & 50.1 & 50.9 & 52.1 & 51.1 & 52.9 & 51.1  \\ 
LLAVA-FT  & NeurIPS2023 & \checkmark & \checkmark &  90.8 & 95.2 & 95.1 & 64.5 & 97.5 & 93.9 & 95.6 & 95.0 & 91.0  \\ 
\rowcolor{light-gray} \textbf{ForenX-S1} & ours & \checkmark & \checkmark &  97.9 & 97.8 & 97.7 & 97.4 & 98.0 & 98.0 & 97.7 & 97.8 & \textcolor{lightgreen}{97.8} \\ 
\rowcolor{light-gray} \textbf{ForenX-S2} & ours & \checkmark & \checkmark & 97.9 & 97.8 & 97.7 & 96.9 & 98.0 & 98.0 & 97.8 & 97.0 & \textcolor{lightblue}{97.6}  \\

\bottomrule
    \end{tabular}
}
\vspace{-0.25 cm}
  \caption{\textbf{Cross-Diffusion-Sources Evaluation on the Genimage Dataset.} The model is trained using SDv1.4 as described in \cite{zhu2024genimage}. LLAVA-PE refers to LLAVA-Prompt-Engineering, while LLAVA-FT denotes LLAVA-FineTuning. We highlight the highest and the second highest numbers in \textcolor{lightgreen}{green} and \textcolor{lightblue}{blue} respectively.}
  \label{tab:SOTA1}
  \vspace{-0.25 cm}
\end{table*}

\begin{table*}[!ht]
    \centering
\resizebox{\textwidth}{33mm}{
    \begin{tabular}{l c c c c c c c c c c c | c}
    \bottomrule \hline
        \multirow{2}*{\textbf{Methods}} & \multirow{2}*{\textbf{Venues}}  & \multicolumn{2}{c}{\textbf{Abilities}} & \multicolumn{9}{c}{\textbf{Test Models}}     \\
          \cmidrule(lr){3-4}\cmidrule(lr){5-13} ~ & ~ &    \textbf{Detection} & \textbf{Explain}  & \textbf{ProGAN} & \textbf{StyleGAN} & \textbf{StyleGAN2} & \textbf{BigGAN} & \textbf{CycleGAN} & \textbf{StarGAN} & \textbf{GauGAN} & \textbf{Deepfake} & \textbf{mAcc}\\
          \bottomrule \hline 
CNNDet\cite{wang2020cnn} & CVPR2020 & \checkmark & $\times$ & 91.4 & 63.8 & 76.4 & 52.9 & 72.7 & 63.8 & 63.9 & 51.7 & 67.1 \\ 
      Frank\cite{Frank}                     & PMLR2020 &  \checkmark & $\times$ & 90.3 & 74.5 & 73.1 & 88.7 & 75.5 & 99.5 & 69.2 & 60.7 & 78.9 \\ 
      Durall\cite{Durall}                   & CVPR2020 &  \checkmark & $\times$ & 81.1 & 54.4 & 66.8 & 60.1 & 69.0 & 98.1 & 61.9 & 50.2 & 67.7\\ 
      Patchfor\cite{chai2020makes}          & ECCV2020 &  \checkmark & $\times$ & 97.8 & 82.6 & 83.6 & 64.7 & 74.5 & 100. & 57.2 & 85.0 & 80.7 \\
      F3Net\cite{qian2020thinking}          & ECCV2020 &  \checkmark & $\times$ & 99.4 & 92.6 & 88.0 & 65.3 & 76.4 & 100. & 58.1 & 63.5 & 80.4 \\
      SelfBland\cite{shiohara2022detecting} & CVPR2022 &  \checkmark & $\times$ & 58.8 & 50.1 & 48.6 & 51.1 & 59.2 & 74.5 & 59.2 & 93.8 & 61.9 \\
      GANDet\cite{mandelli2022detecting}    & ICIP2022 &  \checkmark & $\times$ & 82.7 & 74.4 & 69.9 & 76.3 & 85.2 & 68.8 & 61.4 & 60.0 & 72.3 \\
      BiHPF\cite{jeong2022bihpf}            & WACV2022 &  \checkmark & $\times$ & 90.7 & 76.9 & 76.2 & 84.9 & 81.9 & 94.4 & 69.5 & 54.4 & 78.6 \\
      FrePGAN\cite{jeong2022frepgan}        & AAAI2022 &  \checkmark & $\times$ & 99.0 & 80.7 & 84.1 & 69.2 & 71.1 & 99.9 & 60.3 & 70.9 & 79.4 \\ 
      LGrad \cite{Tan2023CVPR}              & CVPR2023 &  \checkmark & $\times$ & 99.9 & 94.8 & 96.0 & 82.9 & 85.3 & 99.6 & 72.4 & 58.0 & 86.1\\
      UniFD \cite{ojha2023towards}          & CVPR2023 &  \checkmark & $\times$ & 99.7 & 89.0 & 83.9 & 90.5 & 87.9 & 91.4 & 89.9 & 80.2 & {{89.1}}\\
      FreqNet\cite{tan2024frequency}         & AAAI2024 &  \checkmark & $\times$ & 99.6 & 90.2 & 88.0 & 90.5 & 95.8 & 85.7 & 93.4 & 88.9 & 91.5\\
      NPR\cite{tan2024rethinking}            & CVPR2024 &  \checkmark & $\times$ & 99.8 & 96.3 & 97.3 & 87.5 & 95.0 & 99.7 & 86.6 & 77.4 & {92.5} \\
      FatFormer\cite{liu2024forgery}         & CVPR2024 &  \checkmark & $\times$ & 99.9 & 97.2 & 98.8 & 99.5 & 99.3 & 99.8 & 99.4 & 93.2 &  \textcolor{lightgreen}{98.4}\\
 \bottomrule
  LLAVA    & NeurIPS2023 & \checkmark & \checkmark & 61.9 & 50.2 & 49.9 & 54.8 & 53.6 & 52.8 & 61.3 & 50.5 &  54.4 \\ 
  LLAVA-PE & NeurIPS2023 &  \checkmark & \checkmark & 63.5 & 50.1 & 50.0 & 54.9 & 59.4 & 53.2 & 59.5 & 51.6 & 55.3 \\ 
  LLAVA-FT & NeurIPS2023 &  \checkmark & \checkmark & 96.7 & 61.6 & 53.9 & 84.8 & 89.6 & 85.0 & 95.3 & 65.1 & 79.0 \\
 \rowcolor{light-gray} ForenX & ours &  \checkmark & \checkmark & 99.9 & 94.8 & 89.8 & 98.6 & 96.6 & 93.2 & 99.1 & 83.2 & \textcolor{lightblue}{94.4}\\ 

\bottomrule
    \end{tabular}
}
\vspace{-0.25 cm}
  \caption{\tft{Cross-GAN-Sources Evaluation on ForenSynths dataset.} We adopt ProGAN as the training source following \cite{liu2024forgery,tan2024rethinking}.}
  \label{tab:SOTA2}
      \vspace{-0.45 cm}
\end{table*}

 \begin{figure*}[ht!]
 \vspace{-0.4 cm}
   \centering
        \includegraphics[width=1.00\textwidth]{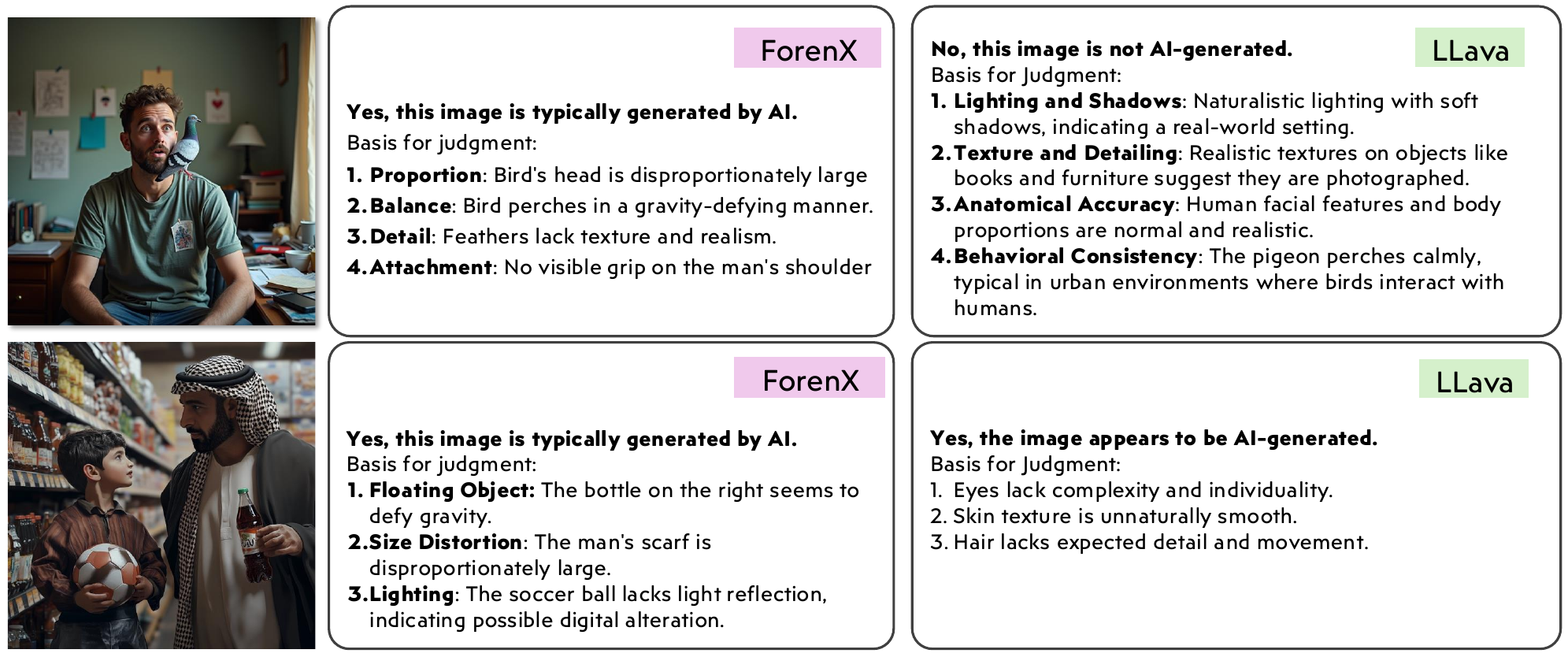}
    \caption{Examples of Explainable AI-Generated Image Detection using LLava and our ForenX. ForenX identifies anomalies in the location and status of objects, such as the pigeon and Coca-Cola, within the images. Due to space limitations, the full description is provided in the supplementary materials.}
    \label{fig:Qualitative-Analysis1}
    \vspace{-0.55 cm}
 \end{figure*}

 \begin{figure}[ht!]
 \vspace{-0.25 cm}
   \centering
    \includegraphics[width=0.45\textwidth]{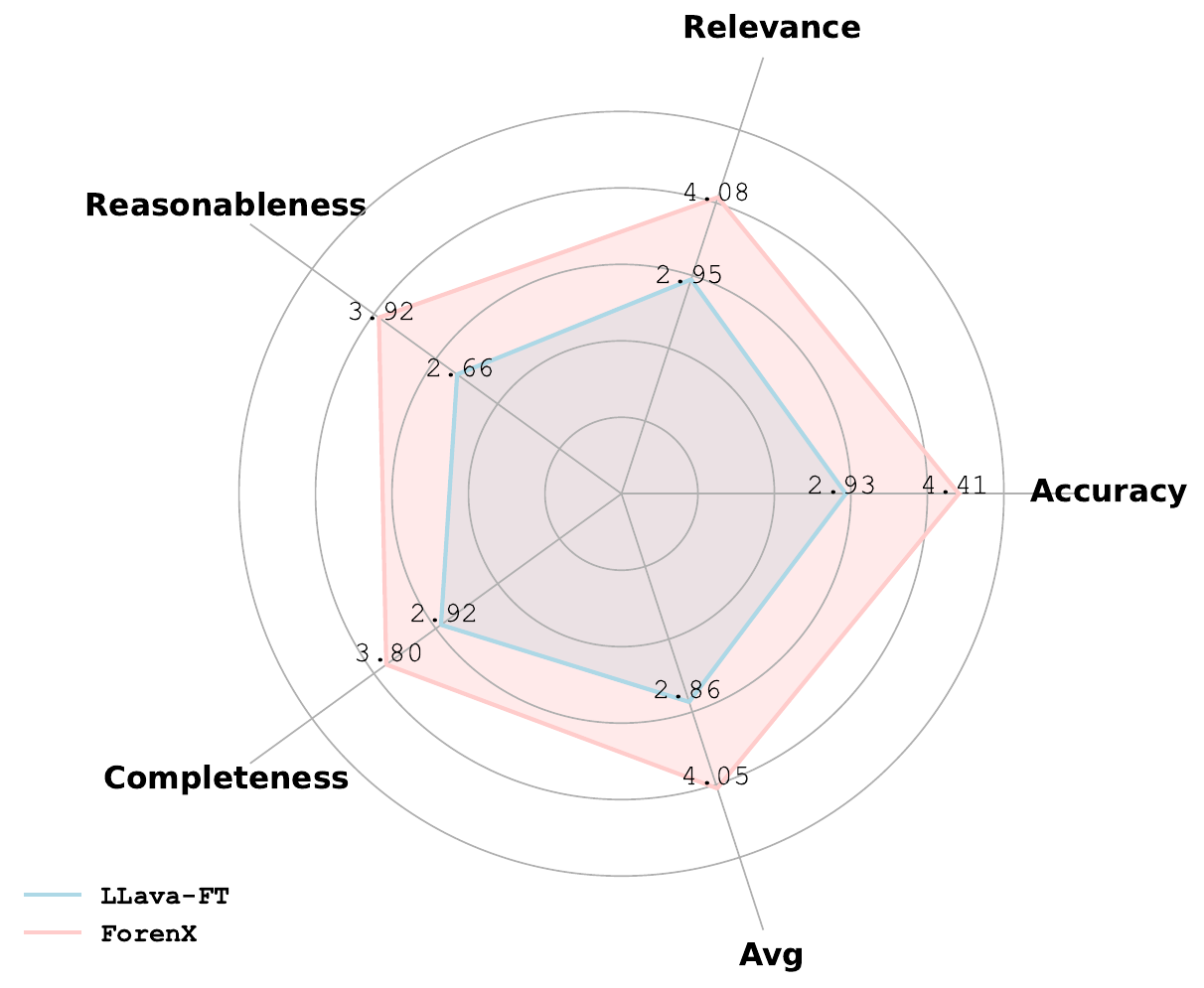}
    \caption{User study for Comparing Explanation Performance Between LLaVA and Our ForenX. We set a full score of 5 points.}
    \label{fig:user_study}
    \vspace{-0.25 cm}
 \end{figure}

\paragraph{Dataset.}
We conduct our experiment on two major benchmarks, \textit{Genimage} and \textit{ForenSynths}, for detection accuracy evaluation and one constructed dataset, \textit{ForgReason}, for explainability fine-tuning.
\begin{itemize}
    \item 
\textbf{Genimage.} This dataset predominantly utilizes various diffusion models for image generation, including Midjourney \cite{Midjourney}, SDv1.4 \cite{rombach2022high}, SDv1.5 \cite{rombach2022high}, ADM \cite{dhariwal2021diffusion}, GLIDE \cite{nichol2021glide}, Wukong \cite{Wukong}, VQDM \cite{gu2022vector}, and BigGAN \cite{BigGAN}. 

\item 
\textbf{ForenSynths.}
Following Wang et al. \cite{wang2020cnn}, our dataset employs ProGAN as part of its training configuration. We utilize a four-class categorization scheme consisting of horse, chair, cat, and car classes as outlined by Tan et al. \cite{tan2024rethinking} and Liu et al. \cite{liu2024forgery}. The test set encompasses eight subsets derived from various generative models: ProGAN \cite{karras2018progressive}, StyleGAN \cite{karras2019style}, BigGAN \cite{BigGAN}, CycleGAN \cite{CycleGAN}, StarGAN \cite{choi2018stargan}, GauGAN \cite{GauGAN}, and Deepfake\footnote{\url{{Deepfake}}}.

\item 
\textbf{ForgReason.} 
This dataset comprises 2215 realistic fake images from Midjourney, alongside 5000 real and 1000 fake sampled from Genimage dataset. 
Each image from Midjourney includes manually annotated fake descriptions. 
All these images collectively for training.

\end{itemize}
\paragraph{Implementation detail.}
Our implementation uses LLava 8B as the foundational Multimodal Large Language Model (MLLM), incorporating Llama3 as the large language model component. We optimize the model using an initial learning rate set to  \(2 \times 10^{-5}\). The batch size is configured to 128, and the model is trained for 1 epoch.
We apply LoRA on the $q\_{proj}$, $k\_{proj}$, and $v\_{proj}$ layers using the Parameter-Efficient Fine-Tuning (PEFT) \cite{peft} library. The hyperparameters for the LoRA layers are set as follows:  $lora\_r = 8$, $lora\_alpha = 16$, and $lora\_dropout = 0.9$. The proposed method is implemented in PyTorch \cite{paszke2019pytorch} and runs on 4 Nvidia A100 GPUs. Consistent with baseline studies \cite{ojha2023towards, liu2024forgery, tan2024rethinking}, we use mean accuracy (mAcc) as the evaluation metric to assess model performance.

\subsection{Recognition and Generalization Evaluation}
We evaluate the recognition and generalization capabilities of ForenX using the GenImage and ForenSynths datasets. 
In the domain of deepfake detection, we believe that the recognition and generalization abilities of the LLM are crucial. The model must have strong detection capabilities and be able to generalize to unknown sources for its forgery explanations to be considered reliable and acceptable.

\noindent
\textbf{GenImage evaluation. } 
To assess the recognition and generalization capabilities of our ForenX model, we use Stable Diffusion 1.4 as the training set and evaluate our method on images generated by eight different models: Midjourney, SDv1.4, SDv1.5, ADM, GLIDE, Wukong, VQDM, and BigGAN. The detection results are presented in ~\cref{tab:SOTA1}, showing that ForenX achieves an impressive accuracy of 97.8\%. During Stage 2 training, employing ForgReason to enhance the model's interpretability results in a slight 0.2\% impact on its recognition performance. Our model outperforms both non-MLLM and MLLM baselines.
Our ForenX is composed of CLIP, LLM, and two projectors, with CLIP primarily used for image feature extraction. Compared to CLIP-based methods such as FatFormer, UnivFD, and CLIP336-lora, ForenX demonstrates a performance improvement of 8-9\%. This indicates that incorporating an LLM enhances the detection performance of CLIP in AI-generated image detection tasks.
Furthermore, DRCT, which uses a pretrained Stable Diffusion model to generate rebuilt images and then performs detection on paired images, achieves lower performance than ForenX. Our method surpasses DRCT by 7.8\%, while also offering the ability to explain the forensics reasons.

In comparison to LLM-based baselines, our approach also shows significant improvements. To evaluate the vanilla LLava model's deepfake detection capabilities, we used prompts similar to those in ForenX. Additionally, by altering prompts to implement prompt engineering, named LLava-PE, we assess LLava's detection ability. Both vanilla LLava and LLava-PE achieved accuracies of only 51.5\% and 51.1\%, respectively. This is due to LLava's design focus on image content understanding, lacking specific insights needed for AI-generated image detection detection. To address dataset bias, we fine-tuned LLava using the same dataset with LoRA. This fine-tuning improved detection performance to 79.7\%, still significantly less than ForenX's 97.8\% accuracy.

Unlike LLava, which exclusively extracts image content for LLM input, our approach extracts both content information and forensics-related details for LLM input, significantly enhancing detection capabilities. 
ForenX-S1  achieves an accuracy of 97.8\%, notably outperforming LLava, LLava-PE, and LLava-finetune by 45.8\%, 46.2\%, and 17.6\%, respectively. This demonstrates that incorporating forensics-relevant information into MLLM can substantially improve detection performance.

\noindent
\textbf{ForenSynths evaluation. } 
We further validate the performance of the proposed method on the ForenSynths dataset, as shown in ~\cref{tab:SOTA2}.
Following the baselines, we adopt four classes setting to train the models, which consist of images of horse, chair, cat, car generated by ProGAN. Then, evaluation on eight generative models containing ProGAN, StyleGAN, BigGAN, CycleGAN, StarGAN, GauGAN and Deepfake.
Compared to the Non-LLM methods, our proposed ForenX approach achieved competitive results with an accuracy of 94.4\%. 
Our method achieves an improvement of 1.9\% compared to the NPR-based approach and a 5.3\% enhancement over the CLIP-only UniFD method.
Our approach achieves a competitive accuracy of 94.4\%, as opposed to the 98.4\% reported by Fatformer. It is important to note that our method possesses both recognition and interpretation capabilities, whereas Fatformer can only detect.
We further compare against LLM-based methods, including LLava, LLava-PE, and LLava-FT.
It can be observed that our approach outperforms these three methods. 

Additionly, to evaluate the effectiveness of our proposed method on state-of-the-art generative models, we conducted experiments using SDV3 and Flux. 
We sample 6,000 images from Stable-Diffusion-3 and Flux.1 [dev], respectively. An equal number of real images are sourced from the GenImage dataset. Our ForenX achieves an accuracy of \textbf{97.7\%} on SDv3 and \textbf{97.8\%} on Flux, respectively.

\subsection{Explainability Analysis}
In this section, we assess the explainability performance of ForenX through qualitative analysis and user-study.

\tft{Quantitative Evaluation}
To evaluate the interpretability of the proposed model, we present an assessment framework utilizing LLMs. Specifically, both the generated forgery explanations and their respective manually annotated references are input into the LLM for evaluation across five metrics: comprehensiveness, relevance, similarity, reasonableness, and average performance.
The annotation method introduced in Sec.~\ref{DataAnnotationwithHuman} is employed to annotate images from Midjoury and Flux. 
ChatGPT-4o is used to compute those metric, with results in Tab.~\ref{tab:interpreta} showing that our proposed method, ForenX, significantly outperforms LLAVA-FT. To mitigate the impact of variability introduced by ChatGPT-4's stochastic nature, we compute the average value across three iterations and use it as the final result.

\begin{table}[!htp]
\centering
\caption{Quantitative results of interpretability (transposed).} 
\begin{tabular}{c|cc}
\hline
Metric           & LLava-FT & ForenX \\ \hline
Comprehensiveness & 80.7     & \textbf{81.2} \\
Relevance         & 71.4     & \textbf{75.0} \\
Similarity        & 60.9     & \textbf{70.5} \\
Reasonableness    & 74.1     & \textbf{77.2} \\
Avg               & 71.8     & \textbf{76.0} \\
\hline
\end{tabular}
\label{tab:interpreta}
\end{table}




\subsubsection{Qualitative analysis}

~\cref{fig:Qualitative-Analysis1} provides the examples of detection explanations given by LLava and \Ours. 
The first image depicts a pigeon standing on a man's shoulder in an indoor scene.  
As shown in the figure, LLava's judgment is incorrect, as it relies on superficial reasoning based on general aspects of the image, such as quality, lighting, and contextual clues, without delivering specific insights related to AI-generated image detection. 
In contrast, \Ours~accurately identifies the image's authenticity and offers detailed, contextually relevant explanations. For instance, \Ours~highlights the unusual presence of a pigeon perched on the man's shoulder, providing a specific and insightful observation pertinent to the image content in a forensic context. 

In addition, in the second image, \Ours~also detects anomalies in the cola. Although LLava recognizes the image as fake, it did not notice the anomalies with the cola bottle and soccer ball.
All explanations from \Ours~are directly tied to the image's content, showcasing its superior capability to interpret and elucidate the reasoning behind detection decisions. This qualitative analysis underscores the enhanced explainability of \Ours~compared to LLava.

\subsubsection{User-Study}

To quantify the explanatory capability of the proposed method, we conducted a user study. Specifically, we selecte 100 realistic-style images from Midjourney and Flux, then rated the reasons for image forgery generated by \Ours~and Llava-FT. We chose 20 users to evaluate five aspects: accuracy, relevance, reasonableness, completeness, and an overall assessment of all aspects. During this process, we randomly shuffle the order of the two reasons so that users could not know which method generated them. The results are shown in ~\cref{fig:user_study}. We can see that our \Ours~outperforms LLava-FT in all four aspects.

To quantify the explanatory capability of the proposed method, we conduct a user study. Specifically, we selecte 100 realistic-style images from Midjourney and Flux, then rate the reasons for image forgery generated by \Ours~and Llava-FT. We chose 20 users to evaluate five aspects: accuracy, relevance, reasonableness, completeness, and overall evaluation of all aspects. We set a full score of 5 points. During this process, we randomly shuffled the order of the two reasons so that users could not know which method was used to generate them. The results are shown in ~\cref{fig:user_study}. It can be seen that our \Ours~outperforms LLava-FT in five aspects simultaneously.
Our method outperforms LLava-FT on five metrics, indicating that the proposed forensic prompt enhances MLLM's recognition ability, and the proposed ForgReason dataset significantly improves MLLM's interpretative ability in low-sample scenarios.

\begin{table}[!ht]
\vspace{-0.15 cm}
    \centering
\resizebox{0.48\textwidth}{14.0mm}{
    \begin{tabular}{c c c c | c}
    \bottomrule   
\makecell{w/o Lora\\on CLIP} & w/o LLM   & \makecell{w/o Forensics\\Projector} & w/o $L_{detction}$ &  mAcc. \\  \hline
     \ding{55}               & \ding{55} &     \ding{55}                       &     \ding{55}      &       84.7    \\ 
    \ding{51}                & \ding{55} &     \ding{55}                       &    \ding{55}       &   85.7    \\ 
    \ding{51}                & \ding{51} &     \ding{55}                       &    \ding{55}       &       91.0    \\
    \ding{51}                & \ding{51} &     \ding{51}                       &    \ding{55}       &       93.7    \\
     \ding{51}               & \ding{51} &     \ding{51}                       &    \ding{51}       &      97.8    \\   
\bottomrule
    \end{tabular} 
}
\vspace{-0.25 cm}
  \caption{{Ablation Study regarding the effectiveness of ForenX's components on Genimage dataset. The results demonstrate an incremental benefit with the inclusion of each module.}  }
  \label{tab:Ablation-Study}
  \vspace{-0.25 cm}
\end{table}

\subsection{Ablation Study}
In this section, we analyze the impact of various components of our ForenX on its performance using the GenImage dataset.
Therefore, we focus on the impact of the following factors: 
1) the effect of $\mathcal{L}_{Detection}$;
2) the effect of forensics projector;
3) the effect of LLM for CLIP, besides providing interpretability;
4) the effect of Lora on CLIP.
When all those module is disable, it is a fixed CLIP(336px) and trainable MLP implementation for detection. 
The results of are shown in ~\cref{tab:Ablation-Study}. 
Compared to non-LLM methods, the addition of LLM effectively improves detection performance. Additionally, $\mathcal{L}_{Detection}$ and the forensics projector introduce an forensic prompt into the LLM, further enhancing detection performance.
The study reveals that each module contributes to overall effectiveness, with improvements observed as additional components are integrated.

\section{Conclusion}
\if false
In this paper, we explore how to empower Multimodal Large Language Models (MLLMs) with 
capabilities to detect AI-generated images
and introduced \Ours, a simple yet effective framework for explainable detection. 
Unlike MLLMs designed for content understanding, such as LLava, which exclusively extracts image content for LLM input, our approach incorporates both content information and forensic prompt, significantly boosting detection capabilities.
Besides, 
we believe that the exploration of other kinds of forensic prompt is a promising direction,
\eg,
frequency features, gradients, and NPR.
We hope that our work will inspire future advancements in utilizing MLLMs for explainable AI-generated image detection.
\fi

In this paper, we explore methods to enhance Multimodal Large Language Models (MLLMs) with capabilities for detecting AI-generated images, introducing \Ours, a simple yet effective framework for explainable detection. Unlike MLLMs designed solely for content understanding, such as LLava, which focus on extracting image content for LLM input, our approach integrates both content information and a forensic prompt, significantly improving detection performance.
Additionally, we believe that further exploration of diverse forensic prompts, such as frequency features, gradients, and NPR\cite{tan2024rethinking}, presents a promising research direction. We hope this work inspires continued advancements in leveraging MLLMs for explainable AI-generated image detection.

\section{Ethics statement}

The AI-generated images employed in this study are exclusively utilized for training a detection model designed to determine whether an image has been generated by artificial intelligence. These datasets will not be repurposed beyond the scope explicitly defined herein. 
To prevent any potential misuse or unethical applications, strict adherence to transparency principles, respect for digital content creators and consumers, as well as compliance with relevant legal frameworks and ethical standards governing AI technologies is maintained throughout this work. By focusing solely on advancing detection technologies, this research seeks to reinforce authenticity and trustworthiness within visual media domains.

\clearpage
\setcounter{page}{1}
\maketitlesupplementary
\renewcommand\thesection{\Alph{section}}
\renewcommand\thefigure{\Alph{figure}}
\renewcommand\thetable{\Alph{table}}
\setcounter{section}{0}
\setcounter{table}{0}
\setcounter{figure}{0}

\noindent
This supplementary material is organized as follows:
\begin{itemize}
    \item In~\cref{Sec:Data_Annotation}, we provide  more detail of label processing of the forgReason dataset.
    \item In~\cref{Sec:Prompt_Engineering}, we discuss the effects of the different prompt.
    \item In~\cref{Sec:qualitative_results}, we present additional qualitative results and discuss limitations in more details.. 
    \item In~\cref{Sec:Ablation_Study_MF}, we provide further ablation studies.
\end{itemize}

\section{Data Annotation}
\label{Sec:Data_Annotation}

In this Section, we show more detail of data annotation. 
To further enhance the forensics interpretability of the LLM, 
  we involve human annotators to label 2,215 images with explanations on why each image is identified as AI-generated. 
The detailed annotation process is as follows:

\vspace{0.3em}
\noindent
\fbox{\parbox{0.98\columnwidth}{
\begin{enumerate}

    \item \textbf{Select Images:} Obtain images from Midjourney~\cite{Midjourney} and choose those that have a realistic style.
    
    \item \textbf{Annotate with boxes}: Manually annotate areas in the images that appear unreasonable using box annotations. Provide a description explaining why these areas are deemed unreasonable.
    
    \item \textbf{Summarize with GPT-4 Vison:} Use GPT-4 Vision to summarize the images and the manual annotations, generating final explanations for why the images are considered AI-generated.

\end{enumerate}
}}
\vspace{0.01em}

\begin{figure*}[ht!]
   \centering
    \includegraphics[width=1.00\textwidth]{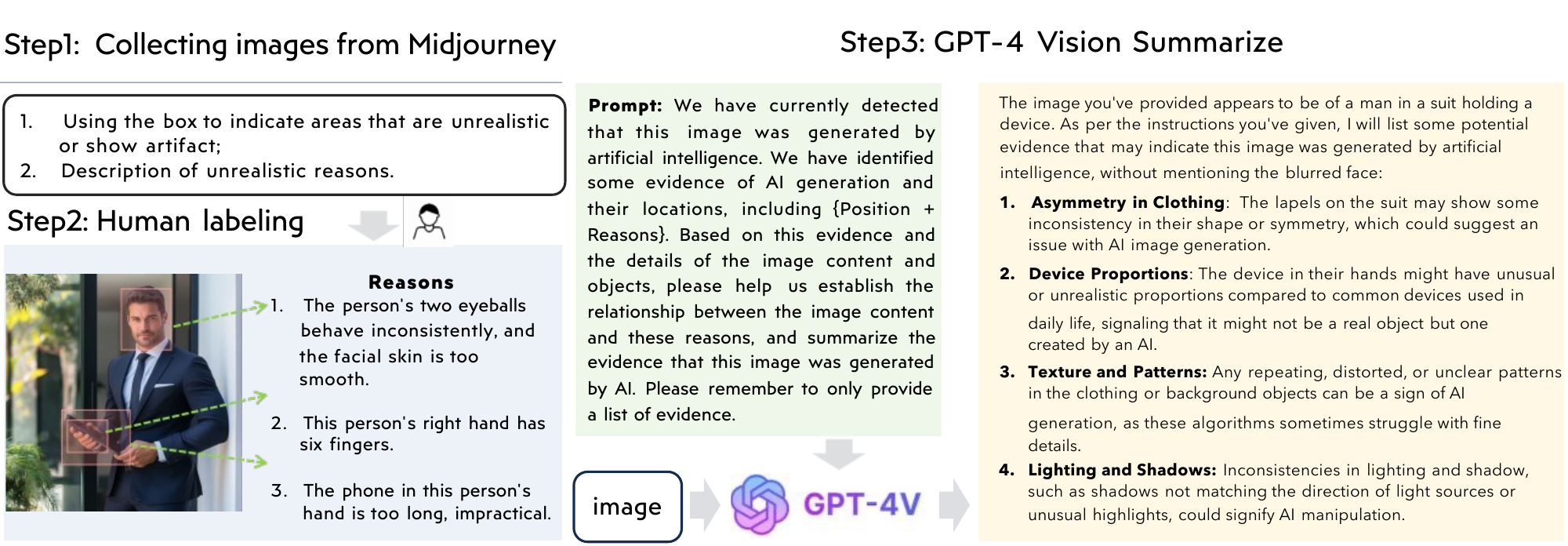}
    \caption{The process of human labeling for AI-generated images involves two steps. First, annotators identify regions with artifacts by marking them with bounding boxes and providing detailed descriptions of the reasons behind their decisions. These location-and-reason pairs are subsequently input into GPT-4 Vision, which generates a summarized annotation for each image.}
    \label{fig:label_processing}
 \end{figure*}

\begin{figure*}[ht!]
   \centering
    \includegraphics[width=1.0\textwidth]{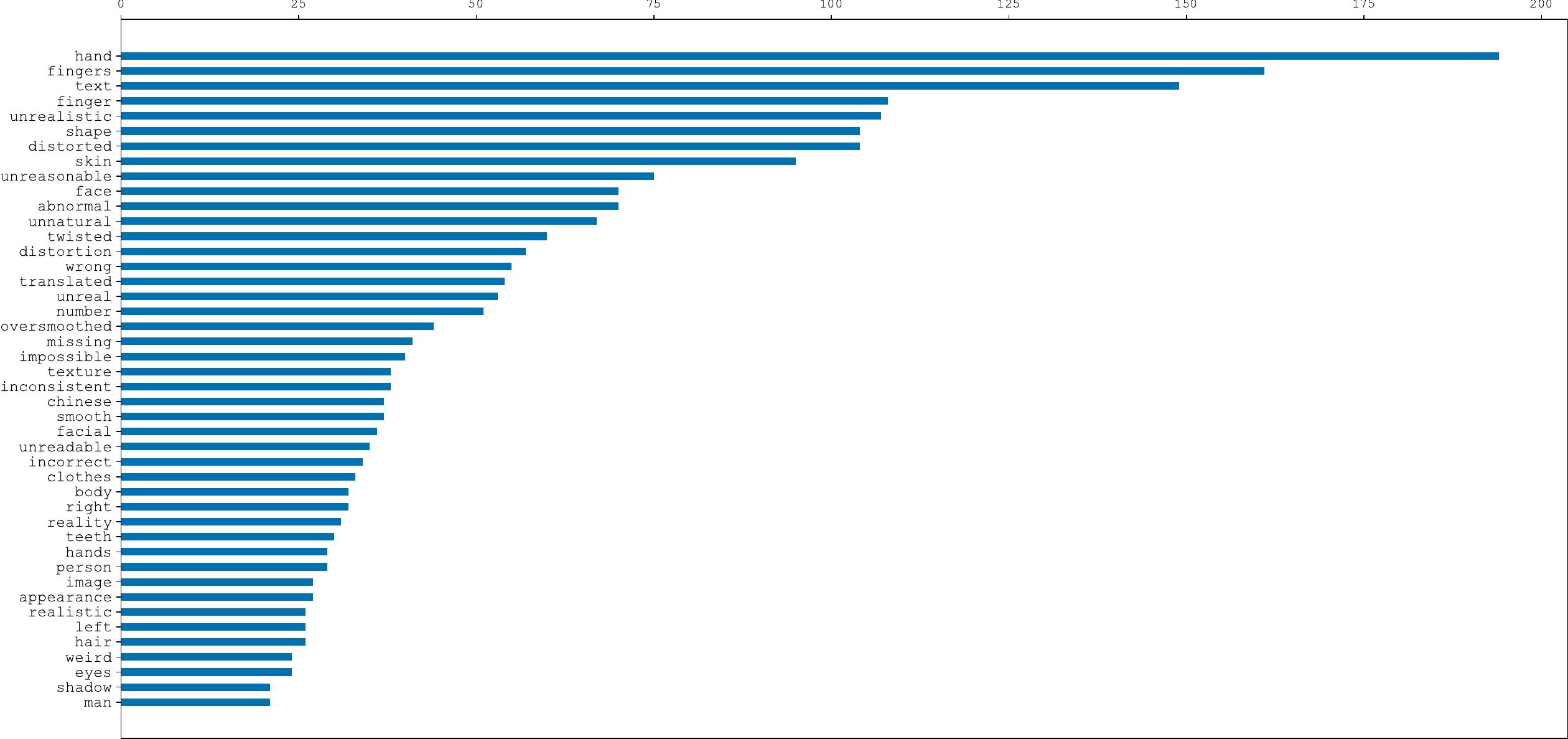}
    \caption{\textbf{Word Frequency Analysis.} The word frequency analysis highlights common forgery region in manual annotation, revealing frequent occurrences of terms associated with human body, including ``hand", ``finger", ``skin", ``teeth", and ``hair". Additionally, prevalent adjectives characterizing forgery include descriptors such as ``unrealistic'', ``distorted'', ``unreasonable'', ``unnatural'', ``twisted'', ``unreal'', ``oversmoothed'', ``impossible,'' and ``unreadable''.}
    \label{fig:word_frequency_analysis}
 \end{figure*}

These images are annotated with captions and detection results following the methods described in Sec. 3.2.1 of main text. 
An example of the annotation process is provided in Figure \ref{fig:label_processing}. 
Before using GPT-4 Vision~\cite{achiam2023gpt} to generate summaries, we convert bounding box coordinates into textual descriptions that convey relative spatial positions. 
In this process, some detailed annotations are omitted to adhere to OpenAI's restrictions. 
We conduct a word frequency analysis of the reasons for manually annotating, as shown in ~\cref{fig:word_frequency_analysis}. 
It can be observed that several words related to the human body appear frequently, such as ``hand", ``finger", ``skin", ``teeth", and ``hair".

\section{Prompt Engineering}
\label{Sec:Prompt_Engineering}

To further investigate the impact of prompts on detection performance, we designed various prompts inspired by prior work\cite{jia2024can}, as follows:
\begin{itemize}[left=1.5em]
\item [v1.] \textbf{System:} \textit{LLava-System-Default} \textbf{User:} Summarize whether this image is Generated by Artificial Intelligence, please return begin with yes or no.

\item [v2.] \textbf{System:} \textit{LLava-System-Default} \textbf{User:} Summarize whether this image is AI-generated images, please return begin with yes or no.

\item [v3.] \textbf{System:} \textit{LLava-System-Default} \textbf{User:} Tell me if there are synthesis artifacts in the image. Please return begin with yes or no.

\item [v4.] \textbf{System:} \textit{LLava-System-Default} \textbf{User:} I want you to work as an image forensic expert for AI-generated image. Check if the image has the artifact. Please return begin with yes or no.

\item [v5.] \textbf{System:} You are an image authentication expert who can identify images as artificially generated by artificial intelligence based on the content and details of the images, and describe it in natural language. \textbf{User:} Summarize whether this image is Generated by Artificial Intelligence, please return begin with yes or no.
\end{itemize}

Here we define the default system prompt of LLava~\cite{llava} as \textit{LLava-System-Default}. We perform prompt engineering evaluation on Geniamge dataset. 
The results are shown in ~\cref{tab:Prompt_engineering}. 
The prompt-v1 is employed in the training stage.  
Despite minor differences between prompt-v1 and prompt-v2, LLava-FT's recognition accuracy exhibit significant variability, decreasing by 8.8\% from an initial rate of 91.0\% to 82.8\%. 
Conversely, our proposed ForenX method demonstrate superior detection capabilities, achieving accuracies of 97.8\% with prompt-v1 and 97.7\% with prompt-v2.
The prompt-v3 and prompt-v4 are designed to enable the large language model (LLM) to determine the presence of artifacts in an image.
Notably, while ForenX and LLava-FT exhibited a decline in performance, LLava's accuracy increased to 58.4\% with the application of Prompt-3. 
Furthermore, we have designed the system prompt to utilize LLM as an image authentication expert within Prompt-v5. 
Notably, the performance of LLava-FT decreased by 2.1\%, dropping from an initial rate of 91.0\% to 87.9\%. 
In contrast, our ForenX achieved a mean accuracy of 97.6\%. 
Our proposed method demonstrates enhanced robustness when responding to various input prompts, compared to existing techniques.

\begin{table*}[!ht]
    \centering
\resizebox{\textwidth}{37mm}{
    \begin{tabular}{l c c c c c c c c c c}
    \bottomrule \hline
        \multirow{2}*{\textbf{Methods}} & \multirow{2}*{\textbf{Prompts}} & \multicolumn{9}{c}{\textbf{Test Models}}     \\
          \cmidrule(lr){3-11} ~ &  ~ & \textbf{Midjourney} & \textbf{SDv1.4} & \textbf{SDv1.5} & \textbf{ADM} & \textbf{GLIDE} & \textbf{Wukong} & \textbf{VQDM} & \textbf{BigGAN} & \textbf{mAcc}\\
          \bottomrule \hline 
LLava     &  \multirow{3}*{ prompt-v1 } & 51.0 & 50.4 & 50.5 & 50.2 & 51.4 & 52.4 & 50.9 & 54.6 & 51.4  \\ 
LLava-FT  &  ~                          & 90.8 & 95.2 & 95.1 & 64.5 & 97.5 & 93.9 & 95.6 & 95.0 & 91.0  \\ 
ForenX    &  ~                          & 97.9 & 97.8 & 97.7 & 97.4 & 98.0 & 98.0 & 97.7 & 97.8 & 97.8\\
\hline 
LLava     &  \multirow{3}*{ prompt-v2 } & 51.1 & 50.3 & 50.3 & 50.1 & 51.0 & 52.1 & 50.3 & 51.9 & 50.9  \\ 
LLava-FT  &  ~                          & 82.2 & 85.8 & 85.7 & 56.9 & 91.5 & 83.3 & 83.7 & 88.4 & 82.2  \\
ForenX    &  ~                          & 97.8 & 97.6 & 97.6 & 97.5 & 97.9 & 97.9 & 97.6 & 97.8 & 97.7 \\ 
\hline 
LLava     &  \multirow{3}*{ prompt-v3 } & 49.5 & 49.5 & 49.2 & 59.9 & 59.5 & 56.9 & 64.1 & 78.2 & 58.4  \\ 
LLava-FT  &  ~                          & 73.6 & 72.7 & 73.9 & 73.7 & 74.3 & 74.0 & 74.0 & 73.4 & 73.7  \\ 
ForenX    &  ~                          & 63.4 & 63.5 & 63.6 & 63.7 & 63.7 & 63.4 & 63.4 & 63.6 & 63.5 \\
\hline 
LLava     &  \multirow{3}*{ prompt-v4 } & 50.6 & 51.0 & 51.0 & 50.0 & 51.6 & 53.2 & 51.0 & 56.4 & 51.9  \\ 
LLava-FT  &  ~                          & 50.0 & 50.0 & 50.1 & 50.1 & 50.1 & 50.0 & 50.1 & 50.1 & 50.1  \\ 
ForenX    &  ~                          & 50.7 & 50.6 & 50.6 & 50.7 & 50.8 & 50.5 & 50.7 & 50.7 & 50.7 \\ 
\hline 
LLava     &  \multirow{3}*{ prompt-v5 } & 50.9 & 50.6 & 50.8 & 50.4 & 52.5 & 53.3 & 52.6 & 56.9 & 52.2  \\ 
LLava-FT  &  ~                          & 86.1 & 89.2 & 88.6 & 71.0 & 93.6 & 89.0 & 91.1 & 94.4 & 87.9  \\  
ForenX    &  ~                          & 97.6 & 97.5 & 97.5 & 97.5 & 97.7 & 97.9 & 97.5 & 97.6 & 97.6 \\
\hline 
\bottomrule
    \end{tabular}
}
  \caption{\textbf{Prompt Engineering Evaluation on the Genimage Dataset.} We design five prompt to evaluate the performance of LLava, LLava-FT, and our ForenX.}
  \label{tab:Prompt_engineering}
\end{table*}

 \begin{figure*}[ht!]
   \centering
    \includegraphics[width=1.00\textwidth]{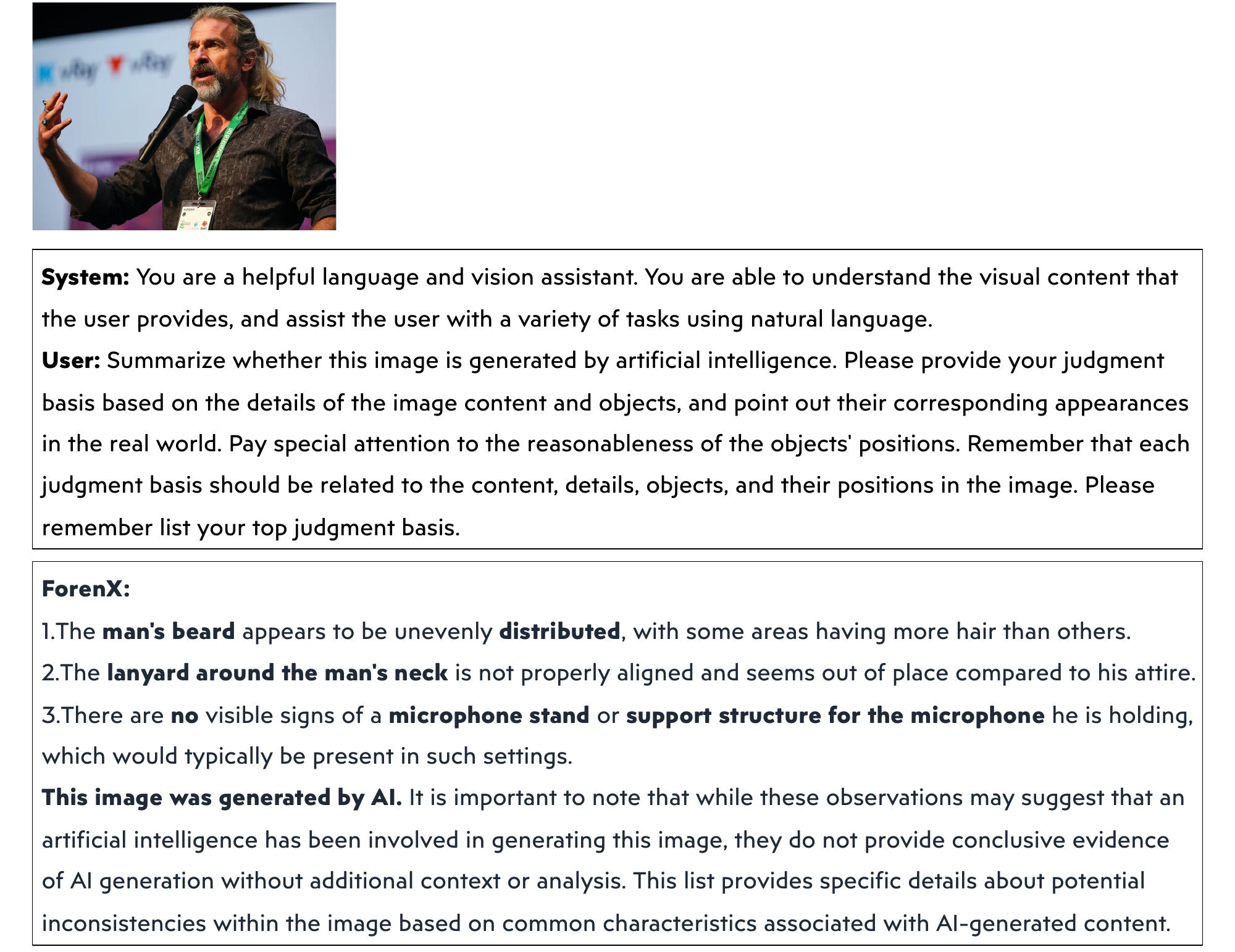}
    \caption{
    \textbf{Qualitative Results Example.} Our ForenX provides three pieces of evidence for detecting AI-generated images by focusing on specific objects, including man's beard, lanyard, and microphone. While humans can readily recognize missing support structures in microphones as evidence, identifying anomalies in a man's beard and lanyard requires closer inspection. In future work, ForenX aims to provide more detailed evidence to assist humans in performing explainable deepfake detection. 
    }
    \label{fig:Qualitative_Results1}
 \end{figure*}

 \begin{figure*}[ht!]
   \centering
    \includegraphics[width=1.00\textwidth]{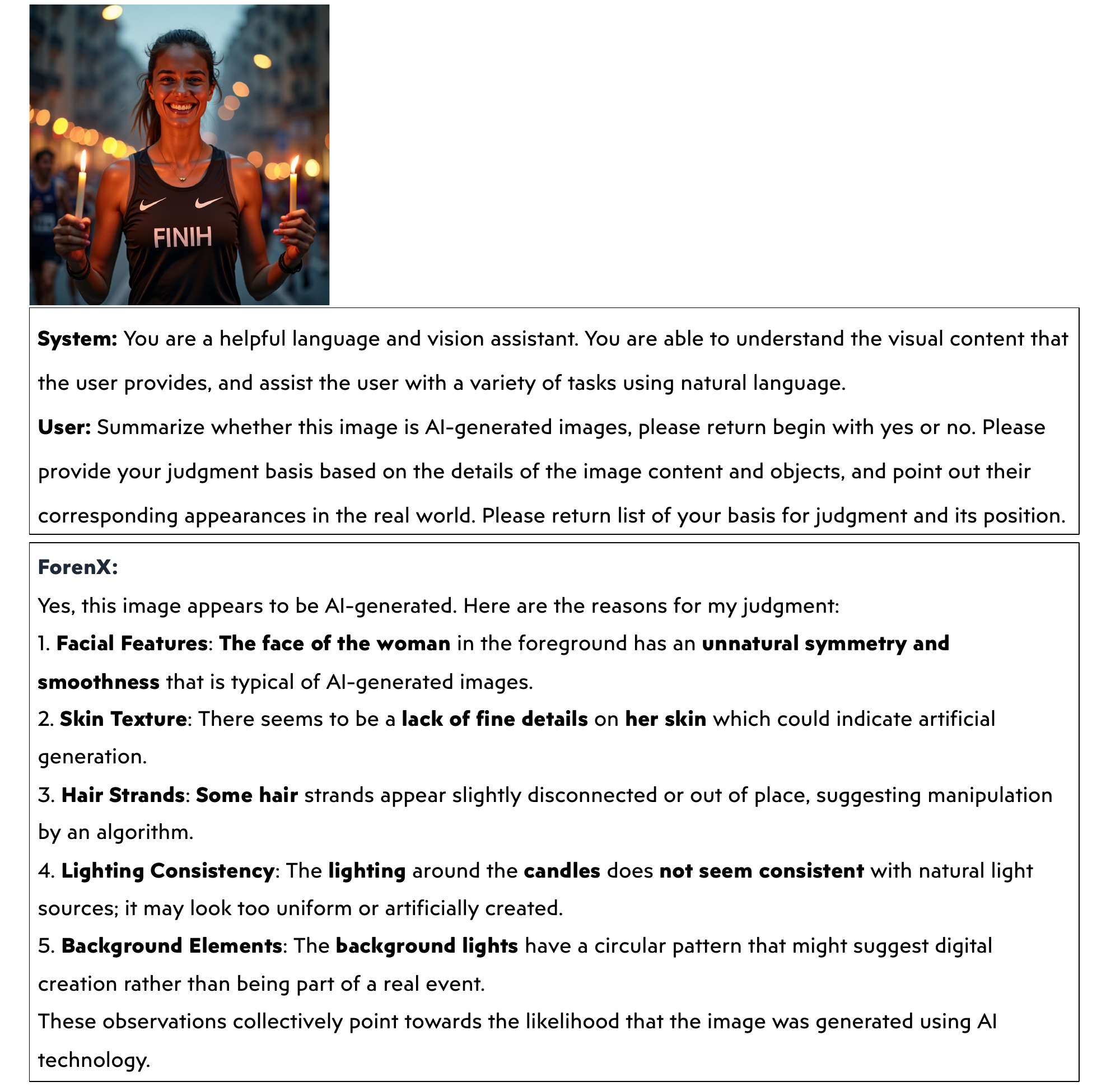}
    \caption{\textbf{Qualitative Results Example.} Our ForenX provides forgery evidences including:  face, skin, hair, candles, and background lights. In those reasons, the characteristics of the woman and candle depicted in the image suggest that it was generated by artificial intelligence; however, the description pertaining to her hair remains ambiguous.   }
    \label{fig:Qualitative_Results2}
 \end{figure*}

 \begin{figure*}[ht!]
   \centering
    \includegraphics[width=1.00\textwidth]{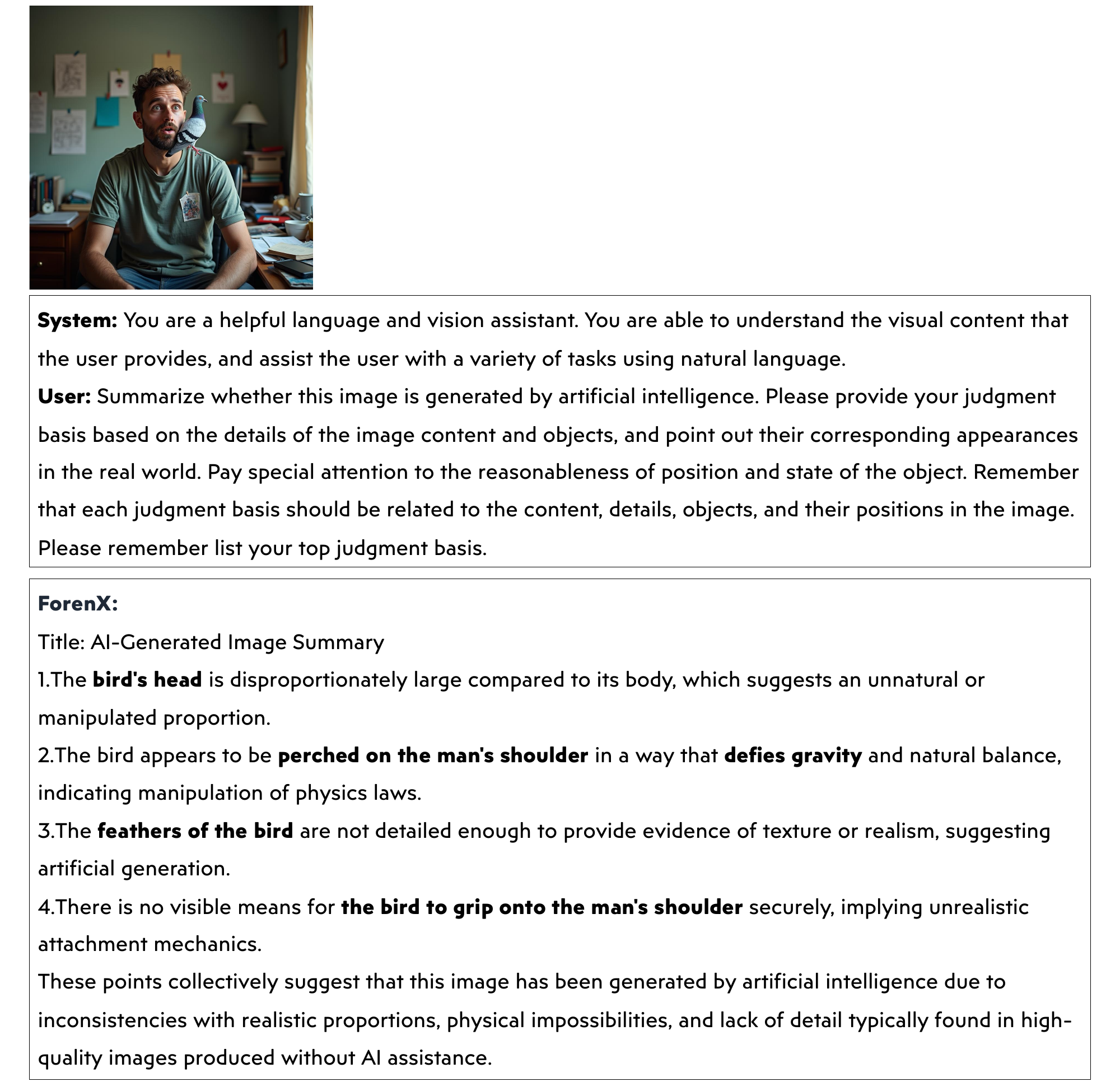}
    \caption{\textbf{Qualitative Results Example.} Our ForenX has twice highlighted the implausibility of pigeons perching on people's shoulders. \textbf{Limitation:} The first reason mentions that the bird's head appears disproportionately large compared to its body; however, additional evidence is required to substantiate this claim.}
    \label{fig:Qualitative_Results3}
 \end{figure*}

 \begin{figure*}[ht!]
   \centering
    \includegraphics[width=1.00\textwidth]{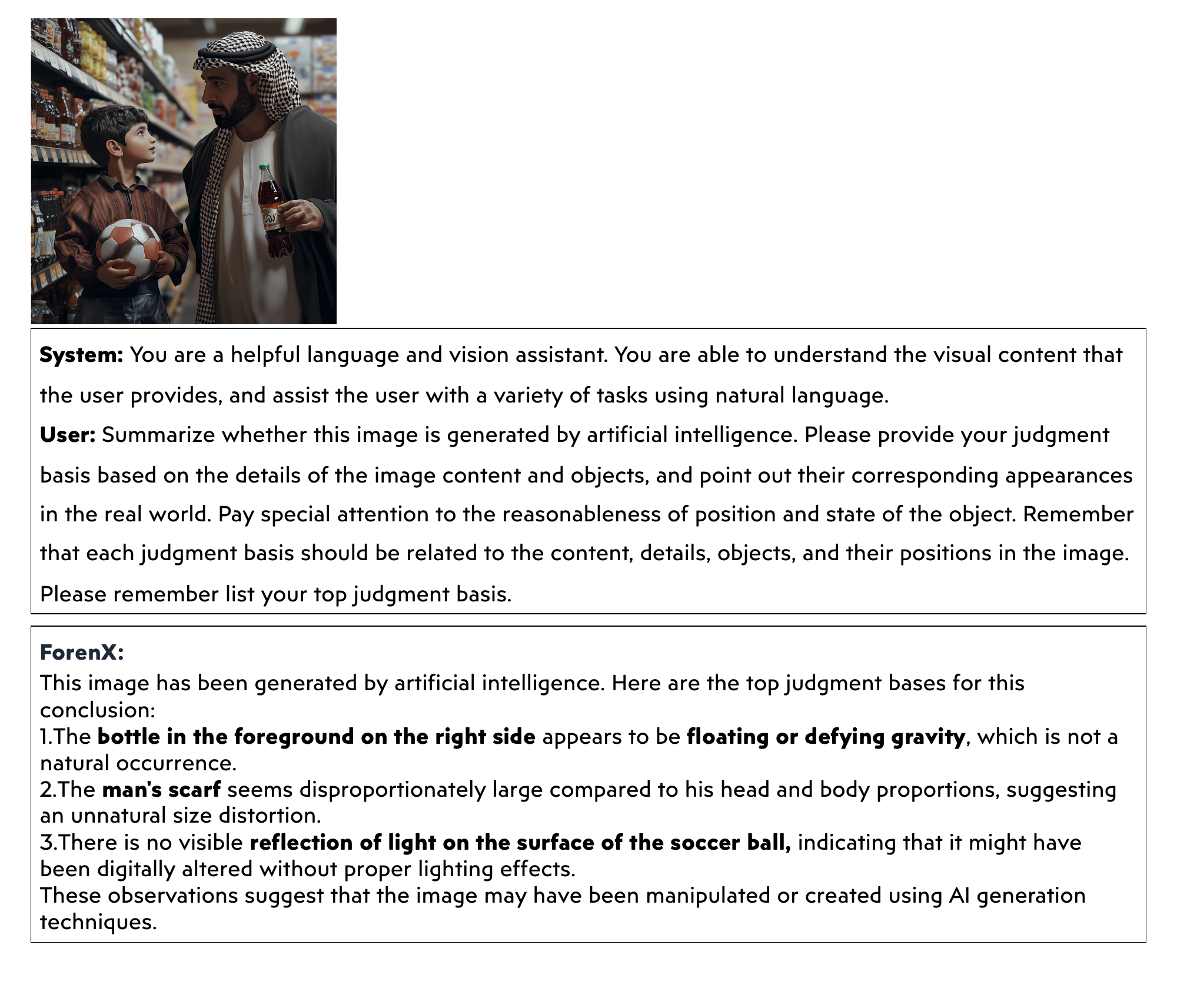}
    \caption{\textbf{Qualitative Results Example.} In our analysis using ForenX, we identify three: bottle, man's scarf, and soccer ball. \textbf{Limitation:} Notably, it overlooks a detail where the liquid contained within the inclined bottle maintains a horizontal surface level. The completeness of forgery reasons needs to be considered in future work.}
    \label{fig:Qualitative_Results4}
 \end{figure*}

 \begin{figure*}[ht!]
   \centering
    \includegraphics[width=1.00\textwidth]{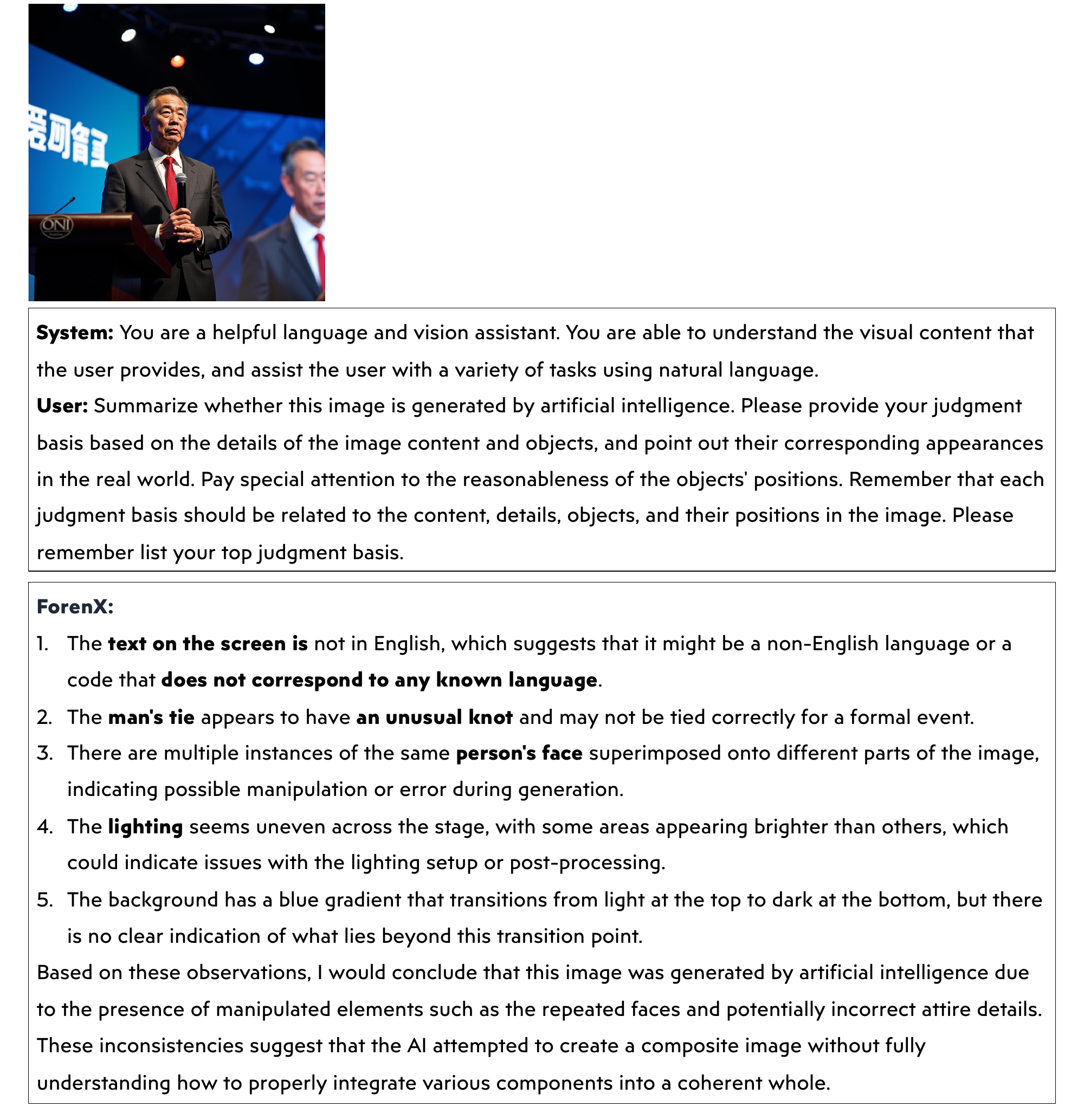}
    \caption{\textbf{Qualitative Results Example}. Our ForenX identifies five distinct indicators of forgery within this image. It identifies that the text on the screen might not belong to any known language and uses this as a basis for forgery detection. \textbf{Limitation:} Furthermore, the description in point four regarding uneven lighting is ambiguous; it requires specification of the affected regions to enhance the persuasiveness of the argument.}
    \label{fig:Qualitative_Results5}
 \end{figure*}

 \begin{figure*}[ht!]
   \centering
    \includegraphics[width=1.00\textwidth]{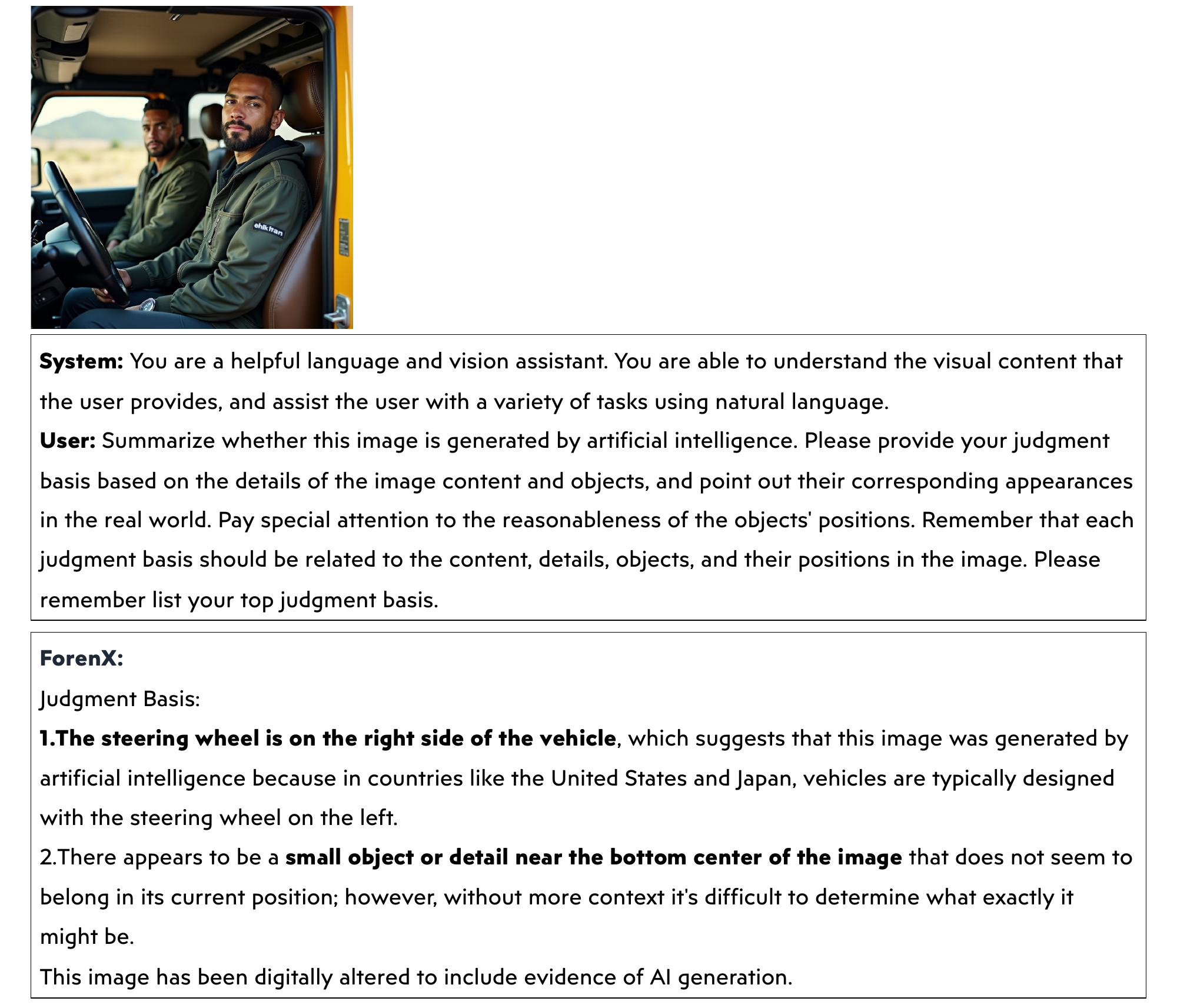}
    \caption{\textbf{Failure Case.} In this image, our ForenX provides incorrect and unintelligible explanations. For instance, it erroneously identifies the steering wheel as being on the right side of the car. Additionally, it notes a small object at the bottom center of the image that cannot be recognized within the picture, which may result from an illusion generated by LLM . }
    \label{fig:Qualitative_Results6}
 \end{figure*}

\section{More Qualitative Results}
\label{Sec:qualitative_results}
Forgery reason generation is considered an open-ended problem within the field. There exists no definitive answer for the reasons behind image forgery. We propose that leveraging large-scale models can generate proposals to assist humans in making informed judgments. 
In this paper, we design a \textbf{Foren}sics e\textbf{X}plainer (\textbf{\Ours}) to identify whether an image is AI-generated and provide reasons. 
The recognition capability of the proposed method has been demonstrated on both the Genimage and ForenSynths datasets. 

Here we further provide an extensive analysis of ForenX by presenting additional qualitative results, focusing on its explainability and limistations. 
{\cblue{Due to space limitations, we place the forgery reason summarized by GPT in the main text. Here we provide the full version of ForenX's forgery reason, as illustrated in ~\cref{fig:Qualitative_Results1}-~\cref{fig:Qualitative_Results4}.}}
In addition, we also provide more qualitative results of ForenX as illustrated in ~\cref{fig:Qualitative_Results5}-~\cref{fig:Qualitative_Results6}.

Our ForenX is capable of identifying causes that are easily discernible by humans and those not visible to the naked eye, while considering possible hallucinations. 
As illustrated in ~\cref{fig:Qualitative_Results1}, ~\cref{fig:Qualitative_Results3}, and ~\cref{fig:Qualitative_Results4}, anomalies such as misplaced microphones, pigeons, and bottles represent issues evident to human observers. Additionally, there exists evidence beyond human perception; for example, in \cref{fig:Qualitative_Results1}, ForenX detects an irregular distribution of a man's beard—an observation challenging for humans without assistance from technology due to its subtlety—and raises questions about whether these observations might result from hallucination effects. 
Additionally, ForenX provides incorrect evidence in ~\cref{fig:Qualitative_Results6}, mistakenly determining that the steering wheel is on the right side and subsequently categorizing this as fake.

\section{Ablation Study}
\label{Sec:Ablation_Study_MF}
In this section, we analyze the impact of detector $f_d$, feature $F_v$, and detection embeding $d$ of our ForenX using the GenImage dataset..

\noindent
\textbf{Detector $f_d$. }
We first analyze the impact of detector $f_d$. 
In our method, detector $f_d$ is utilized to introduce forensics information into $F_f^v$ though an auxiliary forgery detection loss. 
We initially employ a simple summation function $sum(·)$ as $f_d$. For comparison, we replace the $sum(·)$ to MLP layer. The results of this substitution are shown in ~\cref{tab:Ablation_study_1}. 
It can be observed that when we adopt the MLP as the detector $f_d$, we achieve a mean accuracy of 97.6\%, whereas the summation function $sum(·)$ yields a slightly higher mean accuracy of 97.8\%. This suggests that while both methods are effective, the simpler summation approach performs marginally better in this context.

\noindent
\textbf{Feature $F_v$. }
We further analyze the impact of $F_v$ in our ForenX on its performance using the GenImage dataset. In our ForenX, the feature from the visual encoder is fed into the Forensics projector to obtain a forensic prompt. 
In our paper, we use the pooler output of CLIP's last hidden state as 
$F_v$. 
This is also used to perform contrastive learning in CLIP.
For comparison, we experiment with using the entire last hidden state of CLIP as $F_v$. 
In this scenario, the channel size of $F_v$ is first converted from 577 to 16, and then it is fed into the Forensics projector to obtain the forensic prompt. The results are shown in \cref{tab:Ablation_study_2}. 
We observe that our ForenX achieves a mean accuracy of 94.5\% when employing the entire last hidden state of CLIP, whereas ForenX using the pooler output achieves a higher accuracy of 97.8\%. 
This suggests that the pooler output of CLIP's last hidden state contains sufficient information for forensic purposes, effectively feeding forensics information to the LLM. This efficiency highlights the adequacy of the pooler output for generating informative forensic prompts.

\noindent
\textbf{Forensics Embeding $d$. }
We then analyze the impact of detection embeding $d$. 
In our ForenX, a trainable detection embeding $d$ is used to transform the visual feature $F_v$ into forensics features $F_v^f$. We implement this using a $1\times1024 
 ~tensor$,  applying it to $F_v$ through the Hadamard product. For comparison, we experiment with using $1024\times1024 ~tensor$ using matrix multiplication. The results are shown in \cref{tab:Ablation_study_3}. 
 We observe that ForenX with the $1024\times1024$ Forensics Embedding achieves only 96.9\% accuracy. This indicates that the simpler $1\times1024$ tensor with Hadamard product is more effective in transforming the visual features into relevant forensic features, potentially due to its ability to retain essential information without introducing unnecessary complexity.

\section{Limitation Discussion}
ForenX effectively identifies forged images and provides explanations to assist humans in explainable AI-Generated image detection. However, some of these explanations have problems, including:
\begin{itemize}
    \item Hallucinations: Reason contains information unrelated to the prompt and image content.
    \item Vague description: Reason includes partially forgery cues but cannot be verified.
    \item Incorrect information: Reasons include information inconsistent with the image content..
\end{itemize}
In future research, it is essential to develop improved forgery reason datasets and implement more robust training strategies to address these challenges, thereby enhancing the ability of LLMs to explain forgeries.

\begin{table}[!ht]
    \centering
    \begin{tabular}{ >{\centering\arraybackslash}p{3cm} | >{\centering\arraybackslash}p{3cm}}
    \bottomrule
Detector $f_d$          &   mAcc  \\  \hline
$sum()$                 &   97.8      \\  \hline
MLP                     &   97.6      \\     
\bottomrule
    \end{tabular} 
  \caption{{Ablation study on Detector $f_d$.}  }
  \label{tab:Ablation_study_1}
\end{table}

\begin{table}[!ht]
    \centering
    \begin{tabular}{>{\centering\arraybackslash}p{3.5cm} | >{\centering\arraybackslash}p{3cm}}
    \bottomrule
$F_v$   & mAcc  \\  \hline
Pooler                 &   97.8      \\  \hline
All                    &   94.5      \\    
\bottomrule
    \end{tabular} 
  \caption{{Ablation study on Feature $F_v$.}  }
  \label{tab:Ablation_study_2}
\end{table}

\begin{table}[!ht]
    \centering
    \begin{tabular}{>{\centering\arraybackslash}p{3.5cm} | >{\centering\arraybackslash}p{3cm}}
    \bottomrule
Forensics  Embeding $d$  & mAcc  \\  \hline
$1\times1024$                &   97.8      \\  \hline
$1024\times1024$             &   96.9      \\     
\bottomrule
    \end{tabular} 
  \caption{{Ablation study on forensics embeding $d$.}  }
  \label{tab:Ablation_study_3}
\end{table}

\clearpage

{
    \small
    \bibliographystyle{ieeenat_fullname}
    \bibliography{main}
}

\end{document}